\documentclass[10pt,twocolumn,letterpaper]{article}
\usepackage{multirow}
\usepackage{cuted}
\usepackage[pagenumbers]{cvpr} 
\usepackage[table,xcdraw]{xcolor}
\usepackage[percent]{overpic} 
\usepackage{graphicx, xcolor}
\newcommand{\modeltag}[2]{%
  \begin{overpic}[width=\linewidth]{#1}
    \put(0,40){\colorbox{black!60}{\scriptsize \textcolor{white}{\textbf{#2}}}}
  \end{overpic}
}
\definecolor{cvprblue}{rgb}{0.21,0.49,0.74}
\usepackage[pagebackref,breaklinks,colorlinks,allcolors=cvprblue]{hyperref}

\def\paperID{37432} 
\def\confName{CVPR}
\def\confYear{2026}

\title{Distributed Image Compression with Multimodal Side Information at Extremely Low Bitrates}

\author{Guojun Xu, Mingyang Zhang, Jianwen Xiang, Cheng Tan, Yanchao Yang, Junwei Zhou$^*$\\
School of Computer Science and Artificial Intelligence,
Wuhan University of Technology\\
{\tt\small guojunxu@whut.edu.cn, junweizhou@msn.com}}

\begin{document}
\maketitle
\begin{abstract}
Distributed Image Compression (DIC) is crucial for multi-view transmission, especially when operating at extremely low bitrates ($<$ 0.1 bpp). Its core challenge is effectively utilizing side information to achieve high-quality reconstruction under strict bitrate budgets. However, existing DIC approaches struggle to exploit global context and object-level details from side information, leading to local blurring and the loss of fine details in the reconstruction. To address these limitations, we propose a Multimodal DIC framework (MDIC), which, for the first time, leverages side information in a multimodal manner into the DIC paradigm, effectively preserving fine-grained local details and enhancing global perceptual quality in reconstructed images. Specifically, we introduce a text-to-image diffusion-based decoder conditioned on textual side information extracted from correlated images to capture shared global semantics. Moreover, we design a feature-mask generator, supervised by a multimodal fine-grained alignment task, to strengthen the exploitation of visual side information. The generated mask serves two purposes: first, it guides the extraction of fine-grained details from losslessly transmitted side information to preserve the semantic consistency of reconstructed details; second, it regulates the extraction of clustered feature representations from the quantized VQ-VAE embeddings, compensating for category information lost under the extreme compression of the primary image. Extensive experiments on the widely used KITTI Stereo and Cityscapes datasets demonstrate that MDIC achieves state-of-the-art perceptual quality at extremely low bitrates.
\end{abstract}

\section{Introduction}
\label{sec:intro}






\begin{figure}[htbp]
\centerline{\includegraphics[width=0.99  \linewidth]{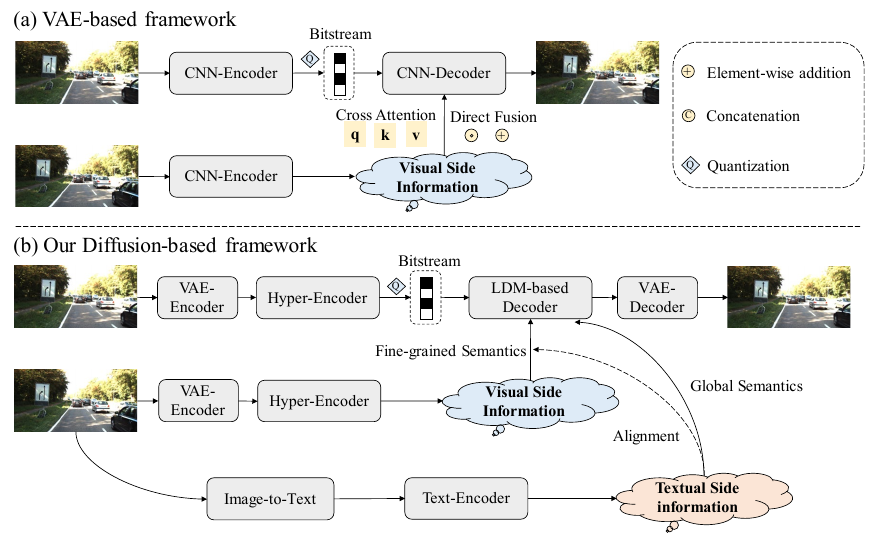}}
\caption{
(a) The existing VAE-based distributed coding framework only with visual side information~\cite{ndic,atn,ldmic,ndsc}. (b) Our diffusion-based pipeline with multimodal side information.}
\label{figure1_2}
\end{figure}



\begingroup
\renewcommand{\thefootnote}{}
\footnotetext{* Junwei Zhou is the corresponding author.}
\endgroup
Distributed Image Compression (DIC) transmits one view losslessly and leverages it as side information to assist in the reconstruction of other compressed views at the decoder~\cite {ndic,ndsc,atn}. In multi-view distributed scenarios such as multi-camera surveillance and 3D scene reconstruction, DIC is particularly important without requiring interaction among encoder devices, theoretically can achieve compression efficiency comparable to joint coding~\cite{wz,sw}. Given their bandwidth-constrained application scenarios and reliance on side information, the DIC algorithms are primarily suited to extremely low bitrates ($<$ 0.1 bits per pixel (bpp)).


The main challenge in DIC lies in effectively utilizing the side information at the decoder, which is highly correlated with the compressed image features~\cite{ndic,atn,ldmic,ref7}. It is crucial to fully exploit informative correlated features while suppressing irrelevant or redundant ones that may interfere with reconstruction. Nevertheless, separating these two types of information is non-trivial due to the complexity and diversity of multi-view object-level details. Moreover, the varying importance of fine-grained information across different regions and objects makes it difficult to balance the model’s attention effectively.

\begin{figure*}[htbp]
\centerline{\includegraphics[width=0.99 \linewidth]{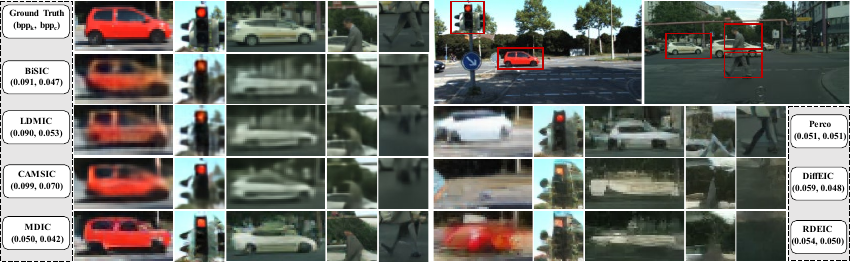}}
\caption{
Comparison of reconstruction performance by joint coding method BiSIC~\cite{bisic} and CAMSIC~\cite{camsic}, the DIC method LDMIC~\cite{ldmic}, the LIC method Perco~\cite {perco}, DiffEIC~\cite{diffeic} and RDEIC~\cite{rdeic}, and the proposed MDIC. $\text{bpp}_k$ and $\text{bpp}_c$ are the compressed bits per pixel (bpp) of these methods in two images. 
}
\label{figurep2}
\end{figure*}

As illustrated in Fig.~\ref{figure1_2}~(a), most existing DIC methods are built upon a VAE framework, where side information is fused through global cross-attention modules~\cite{atn,ldmic,ndsc,ref1}, enabling interaction between compressed features and the auxiliary view. However, under extremely low bitrates, the transmitted features are highly degraded and contain far less information than the side input. As a result, the attention mechanism can focus only on signals correlated with the limited transmitted content, making it difficult to recover lost object-level details during compression. Moreover, these methods prioritize pixel-wise fidelity between reconstructed and original images~\cite{dsin,ndic,ldmic,atn,ref1}, while neglecting the perceptual quality of the reconstruction. When detailed information is missing, they tend to compensate by averaging correlated cues rather than efficiently utilizing side information aligned with the compressed content’s detailed semantics. At extremely low bitrates, this results in excessively smooth reconstructions, exhibiting local blurring, as shown in Fig.~\ref{figurep2}. Representative multi-view coding frameworks such as LDMIC~\cite{ldmic} and BiSIC~\cite{bisic} rely heavily on attention mechanisms but insufficiently exploit fine-grained side information, resulting in blurred textures and degraded perceptual quality.

Recently, perception-oriented compression frameworks~\cite{cdc,ref2,ref3,diffeic,perco,2a,2b} have achieved remarkable progress in Learned Image Compression (LIC), generating high-quality reconstructions from limited information under semantic guidance. Leveraging large-scale pre-training, these models deliver strong perceptual fidelity on benchmark datasets~\cite{mscoco,ref5,ref6}. However, these methods fail to achieve satisfactory performance when applied to distributed scenarios due to the limited scale of trainable multi-view datasets and the absence of dedicated side-information modeling. As illustrated in Fig.~\ref{figurep2}, Perco~\cite{perco}, DiffEIC~\cite{diffeic}, and RDEIC~\cite{rdeic} are diffusion-based LIC framework. Although it yields visually appealing reconstructions, it often produces details inconsistent with the original image and suffers from local distortions.

To address these challenges, we propose a Multimodal Distributed Image Compression (MDIC) framework built upon a pre-trained Latent Diffusion Model (LDM), as illustrated in Fig.~\ref{figure1_2}~(b), where side information is leveraged in a multimodal form. During training, a feature-mask generator, supervised by a multimodal alignment task, is introduced to produce a region-aware mask that adaptively regulates the use of side information. The multimodal alignment provides object-level semantic supervision, guiding the visual mask to capture fine-grained cues with varying semantic importance.


We further incorporate VQ-VAE quantization~\cite{vqvae} to cluster the core category representations. The generated visual mask is applied to the quantized side information to recover category information that is lost during extreme compression. On the other hand, it is applied to the unquantized side information to preserve fine-grained details around key object regions while constraining the reconstructed image to remain semantically consistent with the original content. As illustrated in Fig.~\ref{figurep2}, the proposed MDIC achieves superior perceptual quality while maintaining comparable pixel-wise fidelity.

The contributions of our work are as follows:

\begin{itemize}
\item We propose a novel Multimodal DIC (MDIC) framework that, for the first time, leverages multimodal side information to guide distributed image reconstruction, enabling richer semantic priors and fine-grained detail preservation at extremely low bitrates.

\item We introduce an object-level text prediction task to supervise the proposed feature-mask generator. By enforcing semantic consistency at the object level, the task guides the mask to function as an information gate, selectively retaining category cues and fine-grained structural details.

\item The proposed MDIC achieves impressive perceptual quality at extremely low bitrates, outperforming both existing DIC methods and perception-oriented LIC approaches across multiple perceptual evaluation metrics.

\end{itemize}



\section{Related Work}
\subsection{Distributed Image Compression}

Distributed Image Compression (DIC) is grounded in the Slepian-Wolf theory~\cite{sw}, which shows that distributed coding can match the efficiency of joint coding, and is later extended to lossy compression by Wyner and Ziv~\cite{wz}.



\begin{figure}[htbp]
  \centering


  \begin{subfigure}{0.45\textwidth}
    \centering
    \includegraphics[width=\linewidth]{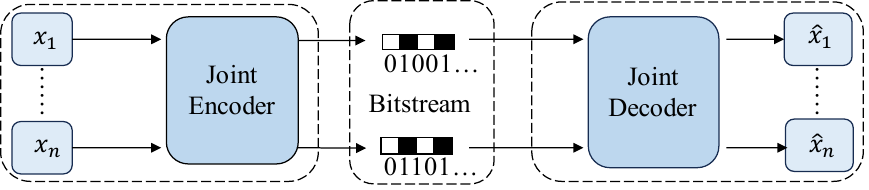}
    \caption{Joint image compression}
    \label{jsc}
  \end{subfigure}

  \vspace{0.5em}

  \begin{subfigure}{0.45\textwidth}
    \centering
    \includegraphics[width=\linewidth]{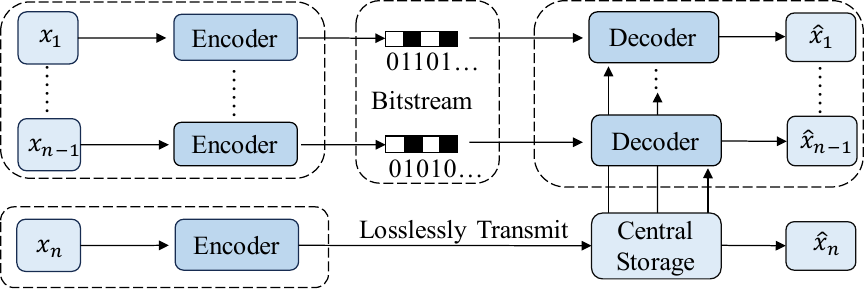}
    \caption{Distributed image compression}
    \label{dsc}
  \end{subfigure}

  \caption{Different compressors for multi-view images.}
  \label{dic_joint}
\end{figure}

As illustrated in Fig.~\ref{dic_joint} (b), learned DIC frameworks predominantly operate under an asymmetric paradigm, where one view is transmitted losslessly to the destination (e.g., central storage or processing unit) and serves as side information for reconstructing other lossy compressed views. The side information incurs no additional bitrate and requires no interaction among encoder-side devices \cite{ndic,ndsc}. Representative methods include NDIC~\cite{ndic}, which extracts shared information from primary and correlated views for decoder-side fusion, and ATN~\cite{atn}, which leverages a cross-attention module to integrate side information during decoding. LDMIC~\cite{ldmic} further extends this idea with a joint context module to better exploit multi-view correlations.

Fig.~\ref{dic_joint}~(a) illustrates a joint coding paradigm, a symmetrical framework in which multi-view images are encoded and decoded jointly, commonly applied to Stereo Image Compression (SIC)~\cite{bssic,camsic}. SASIC~\cite{sasic} uses a stereo attention module to connect left and right views during decoding. ECSIC~\cite{ecsic} explicitly models the mutual information shared across stereo images to improve coding efficiency. BiSIC~\cite{bisic} employs 3D convolutions and bidirectional attention to extract both local and global contextual features.

Despite significant progress in improving distortion metrics such as PSNR~\cite{psnr} and MS-SSIM~\cite{msssim}, these methods suffer from regional and even global blurring at extremely low bitrates. This is because they focus on pixel-wise reconstruction fidelity while neglecting alignment between the distributions of the reconstructed and original images, thereby underutilizing distribution-consistent cues in the side information. In contrast, our approach leverages multimodal side information, in which textual representations capture global distributional information between multi-view images. Benefiting from the multimodal integration and detail-rich generation capability of the pre-trained LDM, our method avoids local pixel averaging and yields sharper reconstructions from highly compressed data.

\subsection{Diffusion-based Compressors}


Diffusion-based image compression methods are mainly designed for extremely low bitrates, aiming to maintain distributional consistency between the original and reconstructed images to achieve high perceptual quality~\cite{diffeic,rdeic,perco}. Perco~\cite{perco} first introduces a pre-trained text-to-image diffusion model to the decoder of the LIC task. DiffEIC~\cite{diffeic} further proposes a two-stage framework that leverages the strong generative capacity of diffusion models for realistic reconstruction at ultra-low bitrates. RDEIC~\cite{rdeic} employs compressed feature initialization and residual diffusion, using noisy latent features to accelerate denoising and improve reconstruction efficiency.

Diffusion-based compressors recover rich global semantics even at extremely low bitrates, naturally benefiting DIC by enabling the separation of global and fine-grained cues within side information. Motivated by this, we propose a DIC framework built upon a pre-trained text-to-image diffusion model, which introduces an object-level multimodal alignment task to generate semantic masks and a mask-gated mechanism to efficiently exploit side information.

\begin{figure*}[htbp]
\centerline{\includegraphics[width=0.99\linewidth]{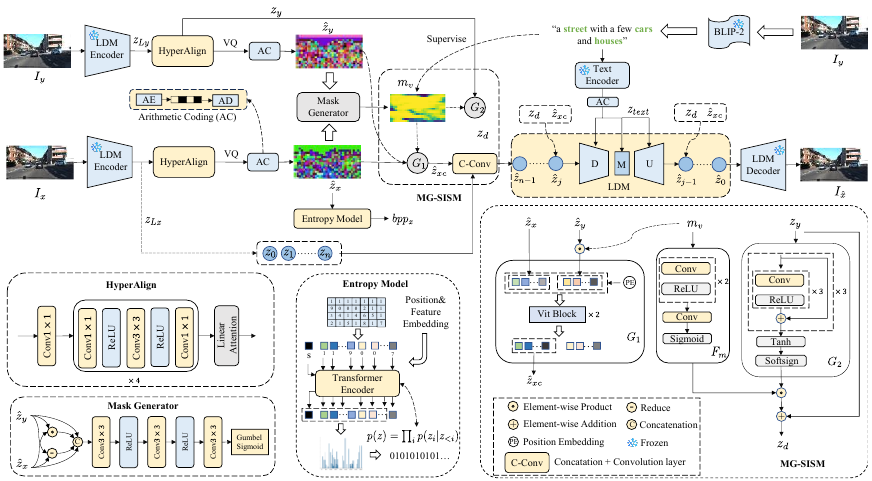}}
\caption{
Overview of the proposed MDIC framework. The input view $I_x$ is lossy-compressed, while the correlated view $I_y$ is transmitted losslessly as side information available only at the decoder. Both $I_x$ and $I_y$ are encoded into latent spaces by the LDM encoder~\cite{sd2}, producing $z_{Lx}$ and $z_{Ly}$, which are further refined by the HyperAlign module into $z_x$ and $z_y$. VQ-VAE quantization~\cite{vqvae} is applied for compression and feature clustering, yielding $\hat{z}_x$ and $\hat{z}_y$. A Transformer-based autoregressive entropy model~\cite{ndsc} estimates the bitrate $bpp_x$. The BLIP-2 captioning model~\cite{blip2} generates textual descriptions from $I_y$, which are then encoded into $z_{text}$ to provide global semantic guidance. The Arithmetic Encoding (AE) further converts the $\hat{z}_x, \hat{z}_y, z_y,$ and $z_{text}$ into a binary bitstream, with the Arithmetic Decoding (AD) performing the inverse operation. Under the supervision of $z_{text}$, a visual mask $m_v$ is generated and integrated into the Mask-Gated Side Information Supplementary Module (MG-SISM): gate $G_1$ fuses $\hat{z}_y$ to restore categorical information lost in $\hat{z}_x$, while gate $G_2$ extracts fine-grained object details from $z_y$. After a denoising process guided by multimodal side information ($z_{d}$, $\hat{z}_{xc}$, and $z_{text}$), the final reconstruction $I_{\hat{x}}$ is obtained by decoding the denoised latent variable $\hat{z}_{0}$ with the LDM decoder.
}
\label{model}
\end{figure*}

\section{Method}
\label{sec:method}

\subsection{Overview of the Proposed Framework}
The overall architecture of the proposed MDIC is illustrated in Fig.~\ref{model}. At the encoder side, a frozen pre-trained VAE encodes the images into a latent space. The encoded features are then refined and compressed using a convolutional and linear-attention hybrid module, producing deep semantic representations for subsequent VQ-VAE quantization and transmission. The overall process is:
\begin{align}
&\begin{cases}
z_{Lx} = E_{vae}(I_x), \\
\hat{z}_x\ \  = VQ\!\left(H_{ax}(z_{Lx})\right),
\end{cases}
\\
&\begin{cases}
z_y = H_{ay}\!\left(E_{vae}(I_y)\right), \\
\hat{z}_y = VQ(z_y),
\end{cases}
\\
&\ \ \ z_{text} = E_{clip}\!\left(B_2(I_y)\right),
\end{align}
where $E_{vae}$ denotes the pre-trained VAE encoder, $VQ$ represents the VQ-VAE quantization module~\cite{vqvae}, $H_{ax}$ and $H_{ay}$ are the HyperAlign module that is similar to Perco~\cite{perco}, $B_2$ refers to the BLIP-2 captioning model~\cite{blip2}, and $E_{clip}$ denotes the CLIP text encoder~\cite{clip}. To estimate the coding cost, a Transformer-based autoregressive entropy model~\cite{ndsc} is employed to model the probability distribution of $\hat{z}_x$, from which the average bitrate $bpp_x$ is derived.

On the decoder side, the Visual Mask Generation Module in Text Supervision (VMGM-TS) generates a visual feature mask to effectively modulate the available visual side information by the Mask-Gated Side Information Supplementary Module (MG-SISM). Then the textual side information $z_{text}$ provides global semantic guidance for the denoising process, while the visual side information focuses on reconstructing fine-grained local details. The combined guidance enables semantically consistent reconstruction, and the final image is obtained through a VAE decoder.




\subsection{Visual Mask Generation Module in Text Supervision (VMGM-TS)
}

\begin{figure}[htbp]
\centerline{\includegraphics[width=0.9\linewidth]{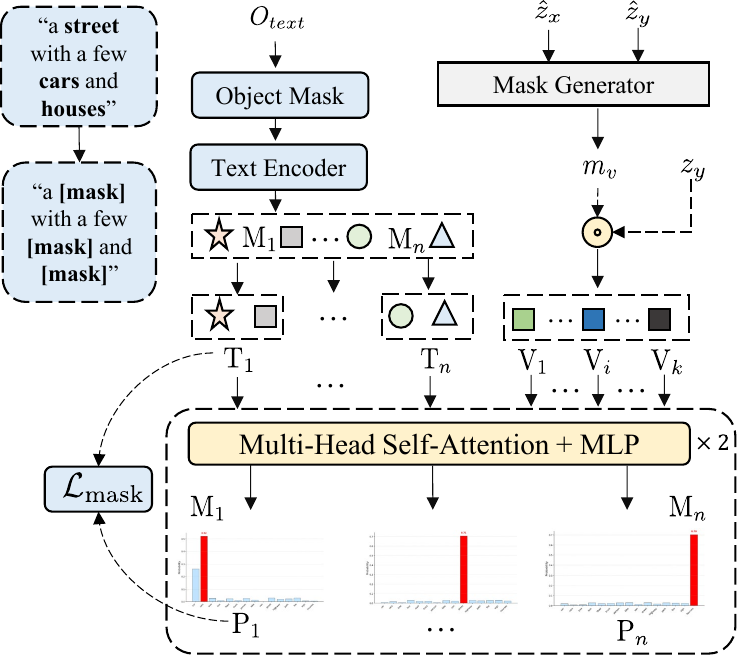}}
\caption{Visual Mask
Generation Module in Text Supervision (VMGM-TS). This module is regarded as an auxiliary task that supervises the Mask Generator, enabling alignment between the object-level visual features and the masked text vocabulary. 
}
\label{VMGM-TS}                             
\end{figure}

To preserve fine-grained details while suppressing redundant multi-view discrepancies, we introduce a Visual Mask Generation Module in Text Supervision (VMGM-TS), as illustrated in Fig.~\ref{VMGM-TS}. The module aligns the generated visual mask with object-level textual semantics, enabling end-to-end learning with visual side information. It identifies spatial regions of target objects and captures their semantic representations, facilitating faithful reconstruction of text-referred content.

To capture fine-grained object semantics, we construct a vocabulary of the $N$ most frequent object nouns in the dataset. During training, object-related words in the image captions are replaced with special mask tokens to form object-masked text. The contextual representations of these masked words are jointly modeled with their corresponding visual masks via an autoregressive-based fusion architecture, enabling alignment between textual context and the visual regions of the same object and facilitating deeper object-level semantic learning beyond explicit word-object correspondences. During the inference phase, the mask can be directly obtained from the trained mask generator.

The overall mask-generating process is as follows:
\begin{gather}
F_{diff} = |\hat{z}_{x} - \hat{z}_{y}|, \quad
F_{prod} = \hat{z}_{x} \odot \hat{z}_{y}, \\
logits = \text{Conv}(\text{concat}(\hat{z}_{x}, \hat{z}_{y}, F_{diff}, F_{prod})), \\
g = -\log(-\log(U + \epsilon) + \epsilon), \; U \sim \mathcal{U}(0,1), \\
m = \sigma\!\left(\frac{logits + g}{\tau}\right), \quad \\
m_v = \mathbb{I}(m > \theta) + sg(m - \mathbb{I}(m > \theta)),
\end{gather}
where $F_{diff}$ and $F_{prod}$ represent the element-wise difference and product capturing disparity and similarity cues. 
The $\text{Conv}(\cdot)$ denotes a three-layer convolutional module with ReLU activation that outputs the values of the mask, as shown in Fig.~\ref{model}. The $g$ is the Gumbel noise sampled~\cite{gambel} from a uniform distribution $\mathcal{U}(0,1)$, and $\tau$ is the temperature controlling the smoothness of the Gumbel–Sigmoid function. The $\theta$ is the hard sampling threshold, and $\mathbb{I}(\cdot)$ is the indicator function for binary masking, $sg(\cdot)$ represents the stop-gradient operation. 

The auxiliary supervision process for the mask generator is as follows:
\begin{gather}
M_{text} = M(O_{text}), \\
\{\text{T}_1, ..., \text{T}_n\} = T_{encoder}(M_{text}), \\
\{\text{V}_1, ..., \text{V}_k\} = P_{token}(m_v \odot z_{y}), \\
\{\text{P}_1, ..., \text{P}_n\} = P_{MSA}(\text{T}_1, ..., \text{T}_n, \text{V}_1, ..., \text{V}_k),
\end{gather}
where $M$ represents the Object Mask. $O_{text}$ is the orange text and $M_{text}$ is the masked text. $P_{token}$ tokenizes the visual features. $T_{encoder}$ encodes the context of masked vocabulary to $\{\text{T}_1,..., \text{T}_n\}$, $n\in{1,2,..., N}$. $\{\text{V}_1, ..., \text{V}_k\}$ are the tokenized visual features obtained by fusing the visual mask $m_v$ with side information $z_y$. $P_{MSA}$ is a mask predictor based on the multi-head self-attention. $\{\text{P}_1, ..., \text{P}_n\}$ is the prediction of the masked text.

\subsection{Mask-Gated Side Information Supplementary Module (MG-SISM)
}

\begin{figure*}[htbp]
\centerline{\includegraphics[width=0.99\linewidth]{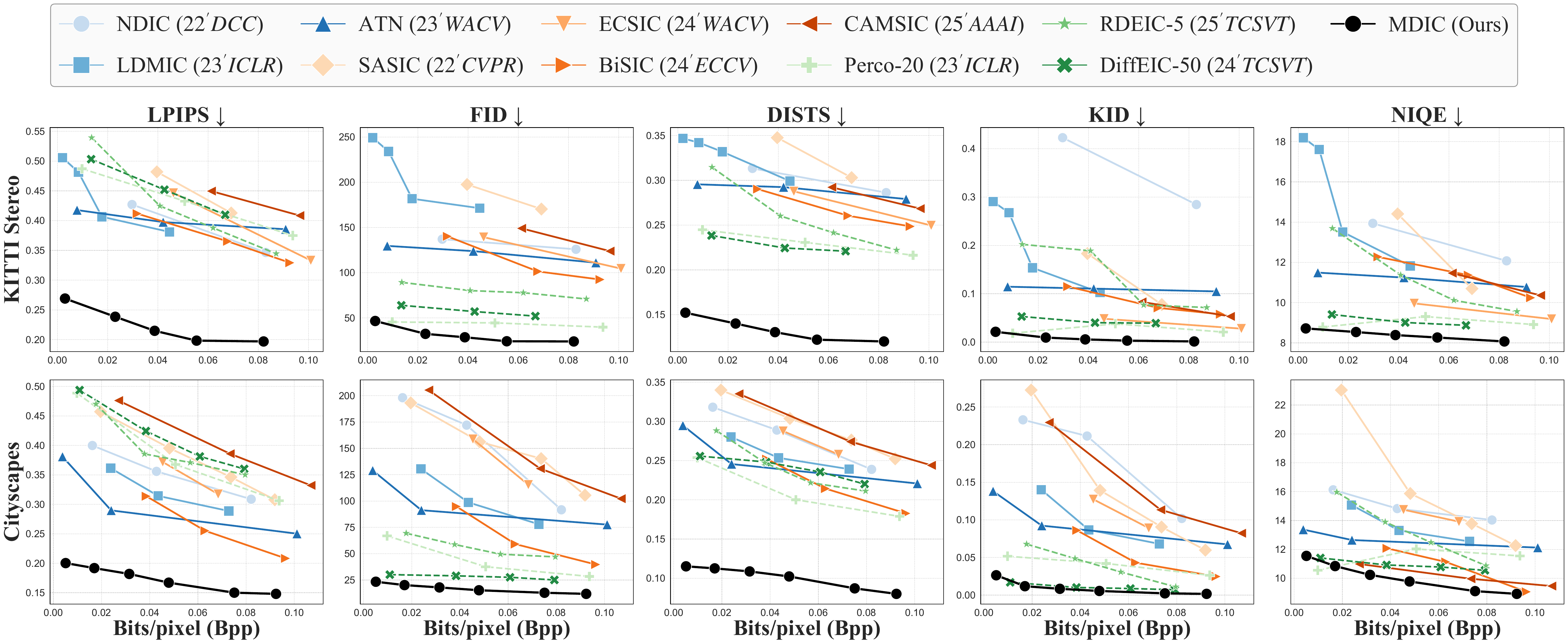}}
\caption{Perception evaluation of MDIC and other DIC, SIC, and LIC methods on KITTI Stereo and Cityscapes datasets. 
}
\label{perception_metric}
\end{figure*}

To further compensate for the loss of categorical information and rich object-level details caused by quantization during compression, we introduce a Mask-Gated Side Information Supplementary Module (MG-SISM). The brief mathematical formulation for MG-SISM can be given as follows:
\begin{gather}
\hat{z}_{xc}  = G_1(\hat{z}_x,\hat{z}_y\odot m_v),\\
z_d  = z_y+G_2(z_y)\odot F_m(m_v).
\end{gather}

As illustrated in Fig.~\ref{model}, VQ-VAE quantization compresses the information while simultaneously clustering image features into a structured latent space, thereby facilitating global reconstruction during denoising. The visual mask $m_v$ is first applied to the quantized side information $\hat{z}_y$ to focus on category-related regions. These masked features interact with the quantized latent representation of the main image through visual Transformer blocks~\cite{vit}, producing semantically enriched category features $\hat{z}_{xc}$.

A gating mechanism further regulates the losslessly transmitted visual side information $z_y$, ensuring that fine-grained object-level details receive maximal attention while suppressing redundant or variant information. This gated output $z_d$ enhances both semantic consistency and perceptual fidelity in the reconstructed images.

\subsection{Decoder with Latent Diffusion Model}
We employ a multimodal-conditioned denoising diffusion model to learn the underlying data distribution, where visual side information provides fine-grained details and overall categorical features, while textual side information guides the denoising toward semantically consistent reconstruction (Fig.~\ref{model}). The diffusion and denoising process is as follows:
\begin{gather}
z_t  = \sqrt{\hat{\alpha}_t} z_{Lx} + \sqrt{1 - \hat{\alpha}_t} \varepsilon, \label{add_noise}\\
q(z_t|z_{t-1})  =\mathcal{N}(z_t;\sqrt{1-\beta_t}z_{t-1},\beta_t I), \label{add_noise1}\\
p_\theta(z_{t-1}|z_t)  =\mathcal{N}(z_{t-1};\mu_\theta(z_t,Z_{cond}, t),\beta_t I). \label{denoise}
\end{gather}

Here, \eqref{add_noise} and \eqref{add_noise1} represent the process of forward noise addition. $\varepsilon \sim \mathcal{N}(0, 1)$ is standard Gaussian noise, $t \in \{1, \dots, T\}$ denotes the diffusion timestep, $\beta_t \in (0,1)$ represents the variance schedule, and $\hat{\alpha}_t = \prod_{i=1}^t (1 - \beta_i)$. The conditional multimodal side information is defined as $Z_{cond} = \{z_{d}, \hat{z}_{xc}, z_{text}\}$,  
where $z_{d}$, $\hat{z}_{xc}$, and $z_{text}$ correspond to fine-grained detail semantics, clustered category features, and global textual information, respectively. During the denoising process, a neural network $\mu_\theta(\cdot)$ predicts the posterior mean based on the noisy input $z_t$, the conditioning information $Z_{cond}$, and the diffusion timestep $t$. 

After the iterative denoising process, the latent representation $\hat{z}_0$ is obtained. 
We then employ a pre-trained VAE decoder $D_{vae}$ to reconstruct the image from the latent space:
\begin{equation}
I_{\hat{x}} = D_{vae}(\hat{z}_0),
\end{equation}
where $I_{\hat{x}}$ denotes the reconstructed image.



\subsection{Loss Function}
We adopt the noise prediction objective as distortion loss:
\begin{equation}
\mathcal{L}_{\text{diff}}
= \mathbb{E}_{z_0,\,\epsilon,\,t}\!
\left[
\| \epsilon - \epsilon_\theta(z_t, t, Z_{cond}) \|_2^2
\right],
\label{eq:noise_prediction_loss}
\end{equation}
where $z_0$ denotes the latent representation initialized from $z_{Lx}$, $\epsilon \sim \mathcal{N}(0, I)$ denotes the Gaussian noise, and $z_t$ is obtained via the forward diffusion process in \eqref{add_noise}.
The function $\epsilon_\theta(\cdot)$ is implemented using a U-Net~\cite{unet} architecture (Fig.~\ref{model}), which consists of a downsampling encoder (D), a middle block (M), and an upsampling decoder (U).

Our main function $\mathcal{L}$ consists of mask auxiliary loss $\mathcal{L_{\text{mask}}}$, vector quantization loss $\mathcal{L_{\text{VQ}}}$, and rate-distortion tradeoff loss $\mathcal{L_{\text{RD}}}$. They are defined as follows:
\begin{gather}
\mathcal{L_{\text{mask}}} = \frac{1}{n} \sum_{i=1}^{n} CE(P_i,T_i),\\
\mathcal{L_{\text{VQ}}} = \mathbb{E}_{z_x} \left[ \| s g(z_x) - \hat{z}_x \|_2^2 + \| s g(\hat{z}_x) - z_x \|_2^2 \right],\\
\mathcal{L_{\text{RD}}}=E_{I_x}[\lambda\cdot E_{\hat{z}_x\sim p(\hat{z}_x\mid I_x)}[-log_{2}({p(\hat{z}_x))}]+\mathcal{L}_{\text{diff}}],\\
\mathcal{L}=\mathcal{L_{\text{VQ}}}+\lambda_{mask}\cdot\mathcal{L_{\text{mask}}}+\mathcal{L_{\text{RD}}},
\end{gather}
where $\mathcal{L}_{\text{mask}}$ supervises the mask generator by measuring the cross-entropy loss between the predicted mask $P_i$ and the ground-truth mask $T_i$ over all $n$ samples. $\mathcal{L}_{\text{VQ}}$ follows the standard VQ-VAE formulation, which consists of a codebook commitment term and a codebook embedding term. The Rate–Distortion loss $\mathcal{L_{\text{RD}}}$ is the core loss of image compression, balancing the trade-off between bitrate and reconstruction quality by hyperparameter $\lambda$. $p(\hat{z}_x)$ is the probability distribution of $\hat{z}_x$, and $\lambda_{mask}$ is set to 0.1.



\section{Experiments}
\subsection{Experimental Settings}
\noindent \textbf{Training Setup and Datasets.} We train and evaluate our model on the KITTI Stereo~\cite{kitti} and Cityscapes~\cite{city} datasets. Following NDIC~\cite{ndic}, 1576 image pairs from KITTI are used for training, and 790 pairs are used for validation and testing. For Cityscapes, we use 2975 image pairs for training, 500 for validation, and 1525 for testing. All KITTI images are center-cropped to $370{\times}740$ and then downsampled to $128{\times}256$, while the Cityscapes images are directly resized to the same resolution.

\begin{figure*}[htbp]
\centerline{\includegraphics[width=0.99\linewidth]{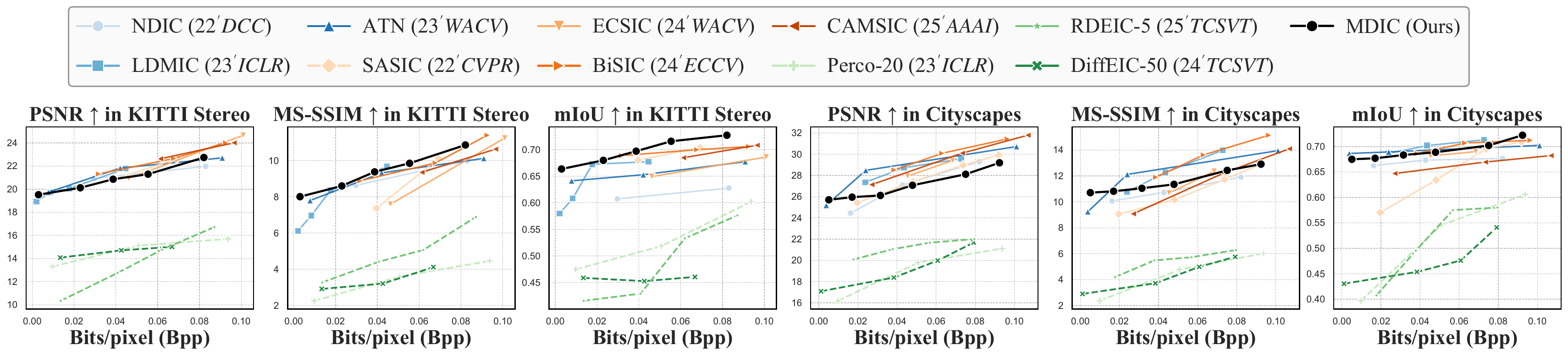}}
\caption{Distortion evaluation of MDIC and other DIC, SIC, and LIC methods on KITTI Stereo and Cityscapes datasets. 
}
\label{distortion_metric}
\end{figure*}

\begin{figure*}[!htb]
  \centering
  \setlength{\tabcolsep}{0.005\textwidth}
  \renewcommand{\arraystretch}{1.1}
  \begin{tabular}{*{6}{>{\centering\arraybackslash}p{0.159\textwidth}}}

    \modeltag{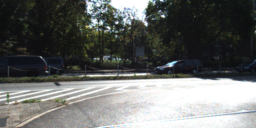}{Ground Truth} &
    \includegraphics[width=\linewidth]{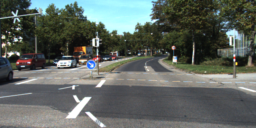} &
    \includegraphics[width=\linewidth]{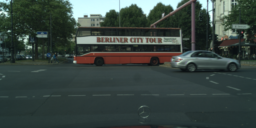} &
    \modeltag{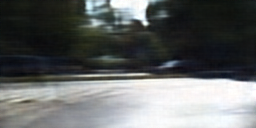}{ATN} &
    \includegraphics[width=\linewidth]{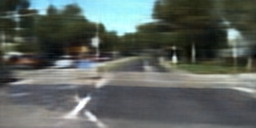} &
    \includegraphics[width=\linewidth]{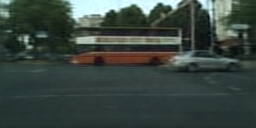} \\[-2pt]

    \makebox[0.159\textwidth]{\scriptsize (Bpp$\downarrow$, PSNR$\uparrow$, LPIPS$\downarrow$)} &
    \makebox[0.159\textwidth]{\scriptsize (Bpp$\downarrow$, PSNR$\uparrow$, LPIPS$\downarrow$)} &
    \makebox[0.159\textwidth]{\scriptsize (Bpp$\downarrow$, PSNR$\uparrow$, LPIPS$\downarrow$)} &
    \makebox[0.159\textwidth]{\scriptsize (0.0316, 23.99, 0.4287)} &
    \makebox[0.159\textwidth]{\scriptsize (0.0401, 22.36, 0.3673)} &
    \makebox[0.159\textwidth]{\scriptsize (0.0248, 28.16, 0.2776)} \\[2pt]

    \modeltag{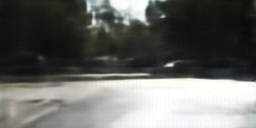}{BiSIC} &
    \includegraphics[width=\linewidth]{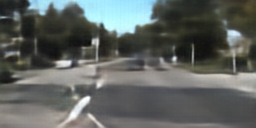} &
    \includegraphics[width=\linewidth]{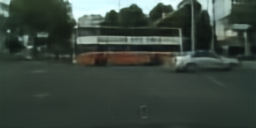} &
    \modeltag{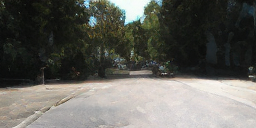}{Perco} &
    \includegraphics[width=\linewidth]{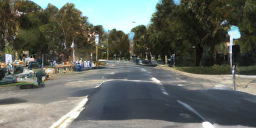} &
    \includegraphics[width=\linewidth]{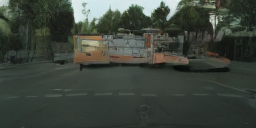} \\[-2pt]

    \makebox[0.159\textwidth]{\scriptsize (0.0288, 23.61, 0.4373)} &
    \makebox[0.159\textwidth]{\scriptsize (0.0408, 21.83, 0.4101)} &
    \makebox[0.159\textwidth]{\scriptsize (0.0460, 27.19, 0.3214)} &
    \makebox[0.159\textwidth]{\scriptsize (0.0313, 17.85, 0.4385)} &
    \makebox[0.159\textwidth]{\scriptsize (0.0313, 15.16, 0.4171)} &
    \makebox[0.159\textwidth]{\scriptsize (0.0313, 18.17, 0.3777)} \\[2pt]

    \modeltag{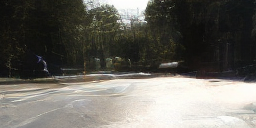}{RDEIC} &
    \includegraphics[width=\linewidth]{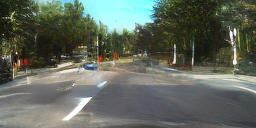} &
    \includegraphics[width=\linewidth]{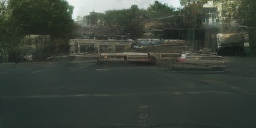} &
    \modeltag{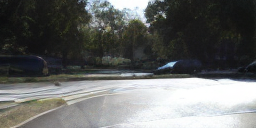}{Ours} &
    \includegraphics[width=\linewidth]{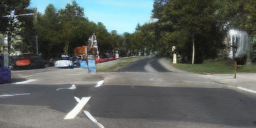} &
    \includegraphics[width=\linewidth]{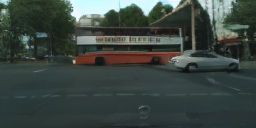} \\[-2pt]

    \makebox[0.159\textwidth]{\scriptsize (0.0188, 12.71, 0.5445)} &
    \makebox[0.159\textwidth]{\scriptsize (0.0212, 12.61, 0.5660)} &
    \makebox[0.159\textwidth]{\scriptsize (0.0425, 20.21, 0.3888)} &
    \makebox[0.159\textwidth]{\scriptsize (0.0107, 22.56, 0.2524)} &
    \makebox[0.159\textwidth]{\scriptsize (0.0201, 20.41, 0.2420)} &
    \makebox[0.159\textwidth]{\scriptsize (0.0207, 24.61, 0.1663)} \\

  \end{tabular}

  \caption{
    Visualization results on KITTI Stereo and Cityscapes datasets.  
    Each row compares two methods (three images per method).  
    Numbers below the images indicate (Bpp, PSNR, LPIPS).  
    Ground Truth includes symbol conventions: $\uparrow$=higher better, $\downarrow$=lower better.
  }
  \label{qual}
\end{figure*}
Our model is trained on four NVIDIA L40 GPUs with a batch size of 8 using the AdamW optimizer. The learning rate follows a constant with a warmup schedule, where it linearly increases to $8{\times}10^{-5}$ during the first 10000 iterations and remains constant thereafter. Training is conducted for an average of 200 epochs per data point. During inference, the number of diffusion steps is set to 10, and the hard sampling threshold $\theta$ is 0.2. We construct the vocabulary consisting of the 14 most frequent object nouns in both datasets. Based on the Rate-Distortion tradeoff setting of Perco~\cite{perco}, we further set the hyperparameter $\lambda$ to \{0.1, 10\}. 

\noindent \textbf{Evaluation Metrics.} We evaluate performance using LPIPS~\cite{lpips}, FID~\cite{fid}, DISTS~\cite{dists}, KID~\cite{kid}, NIQE~\cite{niqe}, as well as PSNR~\cite{psnr} and MS-SSIM~\cite{msssim}. To assess the object-level semantic consistency, we further employ the mean Intersection over Union (mIoU) metric~\cite{miou}.
PSNR and MS-SSIM serve as the primary distortion metrics in DIC, whereas LPIPS, FID, KID, NIQE, and DISTS are commonly used to assess the perceptual realism. Since low-level distortion metrics become less meaningful at extremely low bitrates~\cite{perco}, and our DIC framework is primarily designed for such low-rate scenarios, perceptual quality metrics are adopted as the main evaluation criteria.

\noindent \textbf{Baselines.} We reimplement NDIC~\cite{ndic}, LDMIC~\cite{ldmic}, and ATN~\cite{atn}, where LDMIC~\cite{ldmic} is commonly used as a representative DIC baseline. For comparison with joint coding approaches, we include SASIC~\cite{sasic}, ECSIC~\cite{ecsic}, CAMSIC~\cite{camsic}, and BiSIC~\cite{bisic}. To further compare with diffusion-based image compression frameworks, we evaluate Perco~\cite{perco}, DiffEIC~\cite{diffeic}, and RDEIC~\cite{rdeic}, among which Perco serves as a baseline in LIC methods based on pre-trained latent diffusion models.

\subsection{Main Results}
\noindent \textbf{Comparisons With State-of-the-art (SOTA) Methods.} We compare the proposed MDIC with SOTA DIC and SIC methods, all of which take multi-view images as input for compression and reconstruction. 
Since our work is the first to introduce a perceptually optimized diffusion-based architecture into DIC, we further include benchmark perception-oriented LIC methods for comparison.

As shown in Fig.~\ref{perception_metric}, MDIC achieves the best performance across all perceptual quality metrics in the KITTI Stereo and Cityscapes datasets, significantly surpassing previous DIC approaches. Even when compared with joint coding schemes in SIC, MDIC demonstrates notable improvements in perceptual quality. Moreover, perception-oriented LIC methods, despite the absence of side information, still perform competitively on most perceptual metrics. This is because perceptual optimization aims to align the distribution of the reconstructed image with that of the original one, rather than ensuring pixel-wise fidelity. To comprehensively assess the effectiveness of our method, we further compare it with existing approaches using distortion metrics. As illustrated in Fig.~\ref{distortion_metric}, our method achieves the highest MS-SSIM and mIoU on the KITTI Stereo dataset. Although MDIC achieves a lower PSNR than distortion-oriented methods, it attains a comparable mIoU to the best of them, demonstrating its effectiveness in preserving object-level details. Our distortion metrics significantly outperform other diffusion-based LIC methods, demonstrating its superior ability to maintain semantic consistency. Moreover, at extremely low bitrates, conventional distortion metrics become less meaningful~\cite{perco}.



\noindent \textbf{Visualize comparisons.} 
As illustrated in Fig.~\ref{qual}, we provide qualitative visual comparisons among various methods. At extremely low bitrates, existing DIC methods, such as ATN, tend to suffer from severe local or even global blurring, which significantly degrades perceptual quality. Although joint coding approaches like BiSIC can partially alleviate this issue, they often introduce noticeable local distortions and artifact patterns. Diffusion-based LIC methods, such as Perco and RDEIC, on the other hand, effectively reduce blurring but frequently produce locally inconsistent semantics due to the lack of fine-grained alignment. In contrast, the proposed MDIC effectively avoids these problems. Benefiting from fine-grained perception and multimodal alignment strategies, our method achieves superior perceptual quality with consistent semantics across both global and local regions.

\subsection{Ablation Studies}
\noindent \textbf{Side information $\hat{z}_{xc}$ and $z_d$.} As shown in Table~\ref{tab:bd_metric_1}, we conduct ablation experiments on two types of image side information. When the fine-grained semantic information $z_{d}$ is removed, the diffusion model tends to generate object details (e.g., color or contour) that are inconsistent with the original image. This leads to a noticeable drop in pixel-level fidelity, while having relatively little effect on perceptual consistency. When the category feature information $\hat{z}_{xc}$ is missing, the diffusion model can only generate content from a limited set of categories. Under extremely low bitrates, the absence of sufficient categories severely affects both global and local reconstruction quality of the generated images.
\begin{table}[htbp]
\centering
\setlength{\tabcolsep}{3pt}
\small
\caption{Ablation study of $\hat{z}_{xc}$ and $z_d$. 
BD-Quality denotes the average changes of various quality metrics compared with the MDIC baseline at identical bitrates. Higher is better ($\uparrow$).}
\begin{tabular}{l|cccc}
\toprule
\multirow{2}{*}{\textbf{Method}} & \multicolumn{4}{c}{\textbf{BD-Quality $\uparrow$}} \\
\cmidrule(lr){2-5}
 & \textbf{LPIPS} & \textbf{DISTS} & \textbf{PSNR} & \textbf{MS-SSIM} \\
\midrule
MDIC w/o ($z_d+\hat{z}_{xc}$) & -0.3550 & -0.1826 & -8.6358 & -6.5115 \\
MDIC w/o $z_d$ & -0.1179 & -0.0489 & -2.8161 & -3.1127 \\
MDIC (Ours) & \textbf{0} & \textbf{0} & \textbf{0} & \textbf{0} \\
\bottomrule
\end{tabular}
\label{tab:bd_metric_1}
\end{table}

\begin{figure}[htbp]
\centerline{\includegraphics[width=0.97\linewidth]{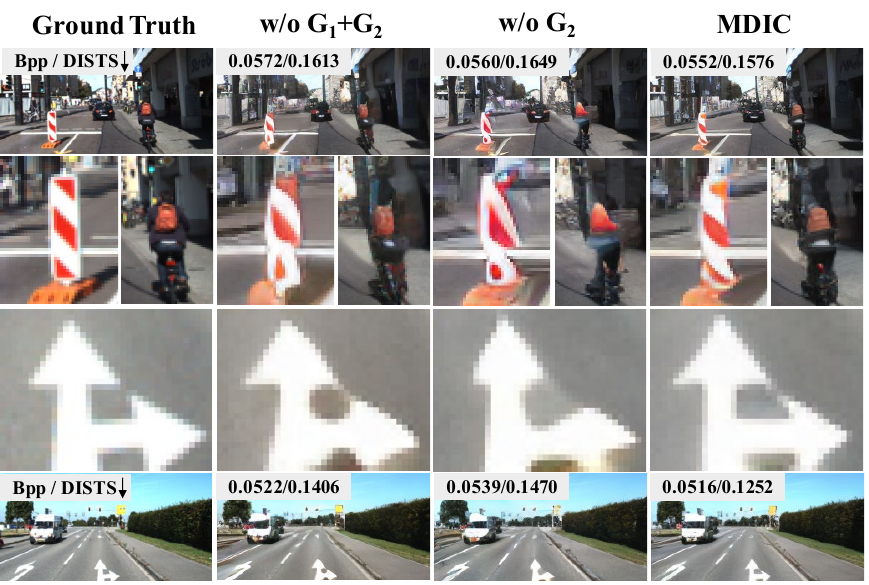}}
\caption{The visual comparison results after ablating the $G_1$ and $G_2$ in MG-SISM. 
}
\label{ablationG}
\end{figure}

\noindent \textbf{The $G_1$ and $G_2$ in MG-SISM.} 
\begin{table}[htbp]
\centering
\setlength{\tabcolsep}{2pt} 
\def\TBLFONT{\small}        
\TBLFONT
\caption{The ablation of the mask-gate $G_1$ and $G_2$. Symbol convention: $\uparrow$=higher better.}
\begin{tabular}{l|cccc}
\toprule
\multirow{2}{*}{\textbf{Method}} & \multicolumn{4}{c}{\textbf{BD-Quality $\uparrow$}} \\
\cmidrule(lr){2-5}
& \textbf{LPIPS} & \textbf{DISTS} & \textbf{PSNR} & \textbf{MS-SSIM} \\
\midrule
MDIC w/o ($G_1+G_2$) & -0.0437 & -0.0050 & -0.5979 & -0.4491 \\
MDIC w/o $G_2$       & -0.0106 & -0.0045 & -0.2864 & -0.2595 \\
MDIC (Ours)       & \textbf{0} & \textbf{0} & \textbf{0} & \textbf{0} \\
\bottomrule
\end{tabular}
\label{tab:bd_metric_2}
\end{table}
As shown in Table~\ref{tab:bd_metric_2}, we perform ablation experiments on the $G_1$ and $G_2$ components within the MG-SISM to validate the effectiveness of our mask-gating mechanism. The results demonstrate that applying mask gating enables the model to better exploit visual side information during reconstruction, while effectively suppressing redundant and inconsistent multi-view features. To further provide an intuitive understanding of the impact of our gating mechanism, 
Fig.~\ref{ablationG} presents visual comparisons between two ablated variants and the MDIC. It is clear that incorporating the gating mechanism facilitates more accurate reconstruction of object boundaries and enhances consistency between category-level and fine-grained semantic details.

Additional experimental results on complexity, hyperparameters, and the role of multimodal side information are provided in the Supplementary Material.

\section{Conclusion}
In this paper, we propose an innovative Multimodal Distributed Image Compression (MDIC) framework that leverages multimodal side information and decodes compressed latent representations with a pre-trained text-to-image diffusion model. A text-supervised visual mask restores category semantics lost in VQ-VAE compression and extracts fine-grained details from visual side information, ensuring semantically faithful reconstruction.
Extensive experiments demonstrate clear perceptual advantages over existing multi-view coding and diffusion-based methods. In the future, we will investigate stronger multi-view correlations and unified pre-trained decoders.

\section*{Acknowledgement} 
This work is partially supported by the Hubei Province Major Science and Technology Innovation Program (2024BAA011), the Key Research and Development Program of Hubei Province (2025BEB012, 2023BAB016).
{
    \small
    \bibliographystyle{ieeenat_fullname}
    \bibliography{main}
}

\clearpage
\setcounter{page}{1}
\maketitlesupplementary



\section{Conditional Entropy Analysis with Multimodal Side Information}

\subsection{Theoretical Analysis}
\noindent \textbf{Theorem.\label{the1}} \textit{Let $X$ be the main source to be compressed, and $Y$ denotes the correlated source. The decoder receives a representation of $Y$ in the form of multimodal features $V = (V_X, V_Y)$ and $T$, where $V_X$ denotes the components of $V$ that are correlated with $X$, and $V_Y$ denotes components specific to $Y$, $T$ represents the textual feature distribution. Then for the joint distribution of $(X,V_X,V_Y,T)$, we have
\begin{equation}\label{eq:main-ineq}
    H(X\mid V_X,V_Y,T)\;\le\; H(X\mid V_X,V_Y).
\end{equation}}

\noindent \textbf{Proof.} Using the chain rule in two different orders, we expand the joint entropy
$H(X,V_X,V_Y,T)$ as
\begin{align}\label{eq1}
&\ \ \ \ \ H(X,V_X,V_Y,T)\notag\\
&= H(V_X,V_Y,T) + H(X\mid V_X,V_Y,T)\\
&= H(V_X,V_Y) + H(T\mid V_X,V_Y) + H(X\mid T,V_X,V_Y)\notag 
\end{align}
\begin{align}\label{eq2}
&\ \ \ \ \ H(X,V_X,V_Y,T)\notag\\
&= H(V_X,V_Y,X) + H(T\mid V_X,V_Y,X) \\
&= H(V_X,V_Y) + H(X\mid V_X,V_Y) + H(T\mid X,V_X,V_Y)\notag 
\end{align}

Equating Eq.~\eqref{eq1} and Eq.~\eqref{eq2}, yielding
\begin{equation} 
\begin{aligned}
H(T\mid V_X,V_Y)+H(X\mid V_X,V_Y,T) \\
= H(X\mid V_X,V_Y)+H(T\mid X,V_X,V_Y).
\end{aligned}
\end{equation} 
Rearranging gives
\begin{equation} 
\begin{aligned}
& \ \ \ \ \ \ H(X\mid V_X,V_Y,T)\\
&= H(X\mid V_X,V_Y)
  - \bigl(H(T\mid V_X,V_Y)-H(T\mid X,V_X,V_Y)\bigr)\\
 &= H(X\mid V_X,V_Y)-I(X;T\mid V_X,V_Y),
\end{aligned}
\end{equation} 
where $I(X;T\mid V_X,V_Y)\ge0$ is the conditional mutual information, thus, we have
\begin{equation} 
\begin{aligned}\label{26}
 H(X\mid V_X,V_Y,T)
 \le H(X\mid V_X,V_Y).
\end{aligned}
\end{equation} 

\noindent \textbf{Remark 1.} Theorem~\ref{the1} and its proof establish that incorporating the additional modality $T$ always reduces or maintains the conditional entropy of $X$ given the decoder side information. The reduction is quantified by the conditional mutual information $I(X; T \mid V_X, V_Y)$, which measures how much information about $X$ is provided by $T$ beyond what is already contained in $V$.   

The minimum achievable coding rate of $X$ with side information $(V,T)$ is
\begin{equation}
\begin{aligned}
&\ \ \ \ \ R^*(V,T) = H(X\mid V_X,V_Y,T) \\
\end{aligned}
\end{equation}

From this expression, $V_Y$ is conditionally independent of $X$ given $(V_X,T)$, which does not help reduce the conditional entropy, thus
\begin{equation}
\begin{aligned}
    H(X\mid V_X,V_Y,T) = H(X\mid V_X,T).
\end{aligned}
\end{equation}

The inequality in Eq.~\ref{26} is strict if and only if $I(X;T \mid V_X,V_Y) > 0$.  
 This occurs when $T$ contains components of $X$ not already captured by $V_X$, or captures complementary semantic information relevant to the reconstruction of $X$. Components in $V_Y$ that are specific to $Y$ and independent of $X$ do not contribute to reducing the conditional entropy. Hence, the compression gain primarily comes from $T$ and $V_X$ that correlate with $X$.
 
\noindent \textbf{Remark 2.} Following Remark~1, identifying the minimum achievable rates as
\begin{equation}
\begin{aligned}
    R^*(V) = H(X \mid V_X), \quad R^*(V,T) = H(X \mid V_X,T).
\end{aligned}
\end{equation}

The expected reduction in rate due to the additional modality $T$ is
\begin{equation}
\begin{aligned}
\Delta R = R^*(V) - R^*(V,T) = I(X;T \mid V_X) \ge 0,\\
\Delta R = 0 
\;\;\Longleftrightarrow\;\;
I(X;T \mid V) = 0
\;\;\Longleftrightarrow\;\;
X \perp T \mid V.
\end{aligned}
\end{equation}

In practice, this implies that multimodal side information can be explicitly leveraged to design more efficient distributed compression schemes. By maximizing $I(X; T\mid V)$ during feature design, one can achieve a significant reduction in the conditional entropy and, consequently, in the required bit rate for encoding $X$. 

\noindent \textbf{Extension to Lossy Compression.} 
The previous analysis assumes lossless compression. In the lossy setting with a distortion constraint $D$, the minimum achievable coding rate is given by the conditional rate-distortion function:
\begin{equation}
\begin{aligned}
R(D \mid V,T) = \min_{p_{\hat X \mid X,V,T}} 
 I(X;\hat X \mid V,T) \ \ 
\text{s.t.}  \   \mathbb{E}[d(X,\hat X)] \le D
\end{aligned}
\end{equation}
where $d(X,\hat X)$ denotes a distortion measure.

Analogously to the lossless case, adding the modality $T$ never increases the minimum achievable rate:
\begin{equation}
R(D \mid V,T) \le R(D \mid V).
\end{equation}

The rate reduction under distortion $D$ is
\begin{equation}
\Delta R(D) = R(D \mid V) - R(D \mid V,T) \ge 0.
\end{equation}

Similar to the lossless case, $\Delta R(D) = 0$ if and only if $X \perp T \mid V$, i.e., $T$ provides no additional information about $X$ given $V$.  

In practice, designing $T$ to maximize its information contribution relative to $V$ helps reduce the rate under a given distortion, enabling more efficient lossy distributed compression.

\subsection{Relative Experiments}
In the proposed MDIC, $T$ is constructed to capture shared multi-view semantic information that complements the visual features $V$. Moreover, $T$ is utilized in the VMGM-TS to supervise the visual feature–mask generator.

\begin{table}[htbp]
\centering
\setlength{\tabcolsep}{3pt}
\small
\caption{Ablation study of the Text Supervision (TS) in VMGM-TS, and textual side information $z_{text}$. BD-Rate measures the average bitrate difference between two rate–distortion (RD) curves over a common quality range, quantifying how much more or less bitrate a method requires, on average, to achieve the same quality level as the baseline. Lower is better ($\downarrow$).}
\begin{tabular}{l|cccc}
\toprule
\multirow{2}{*}{\textbf{Method}} & \multicolumn{4}{c}{\textbf{BD-Rate $\downarrow$}} \\
\cmidrule(lr){2-5}
 & \textbf{LPIPS} & \textbf{DISTS} & \textbf{PSNR} & \textbf{MS-SSIM} \\
\midrule
MDIC w/o TS & 121.49\% & 72.33\% & 97.20\% & 50.78\% \\
MDIC w/o $z_{text}$ & 104.79\% & 82.91\% & 35.67\% & 37.61\% \\
MDIC (Ours) & \textbf{0} & \textbf{0} & \textbf{0} & \textbf{0} \\
\bottomrule
\end{tabular}
\label{tsi}
\end{table}

To demonstrate the impact of textual side information $z_{text}$ on the final reconstruction results, we conduct an ablation study on both $z_{text}$ and Text Supervision (TS) in VMGM-TS. As shown in Table~\ref{tsi}, taking our MDIC as the baseline, removing $z_{text}$ results in higher bitrate 
consumption to achieve the same reconstruction quality. In other words, 
both perceptual metrics (LPIPS, DISTS) and distortion metrics (PSNR, 
MS-SSIM) degrade without $z_{text}$. 

\begin{figure}[htbp]
\centerline{\includegraphics[width=0.97\linewidth]{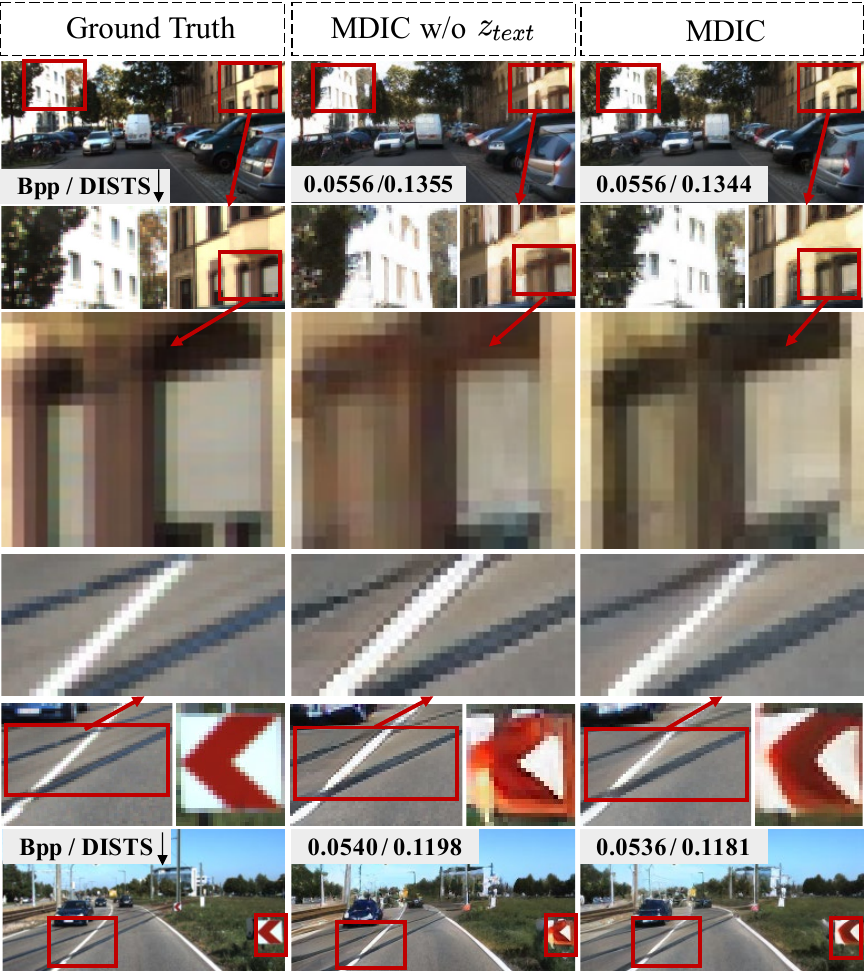}}
\caption{The visual comparison between MDIC and its variant without the textual side information $z_{text}$.}
\label{ablationTSI}
\end{figure}
\begin{figure}[htbp]
\centerline{\includegraphics[width=0.98\linewidth]{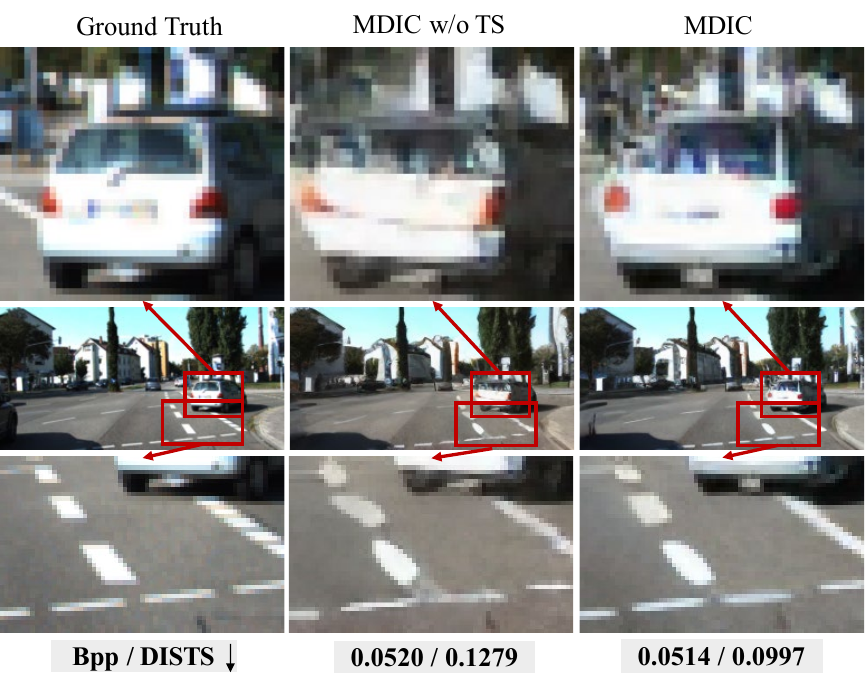}}
\caption{The visual comparison between MDIC and its variant without the Text Supervision (TS) auxiliary task.}
\label{ablationTS}
\end{figure}

\begin{figure}[htbp]
  \centering
  \begin{subfigure}{0.49\linewidth}
    \centering
    \includegraphics[width=\linewidth]{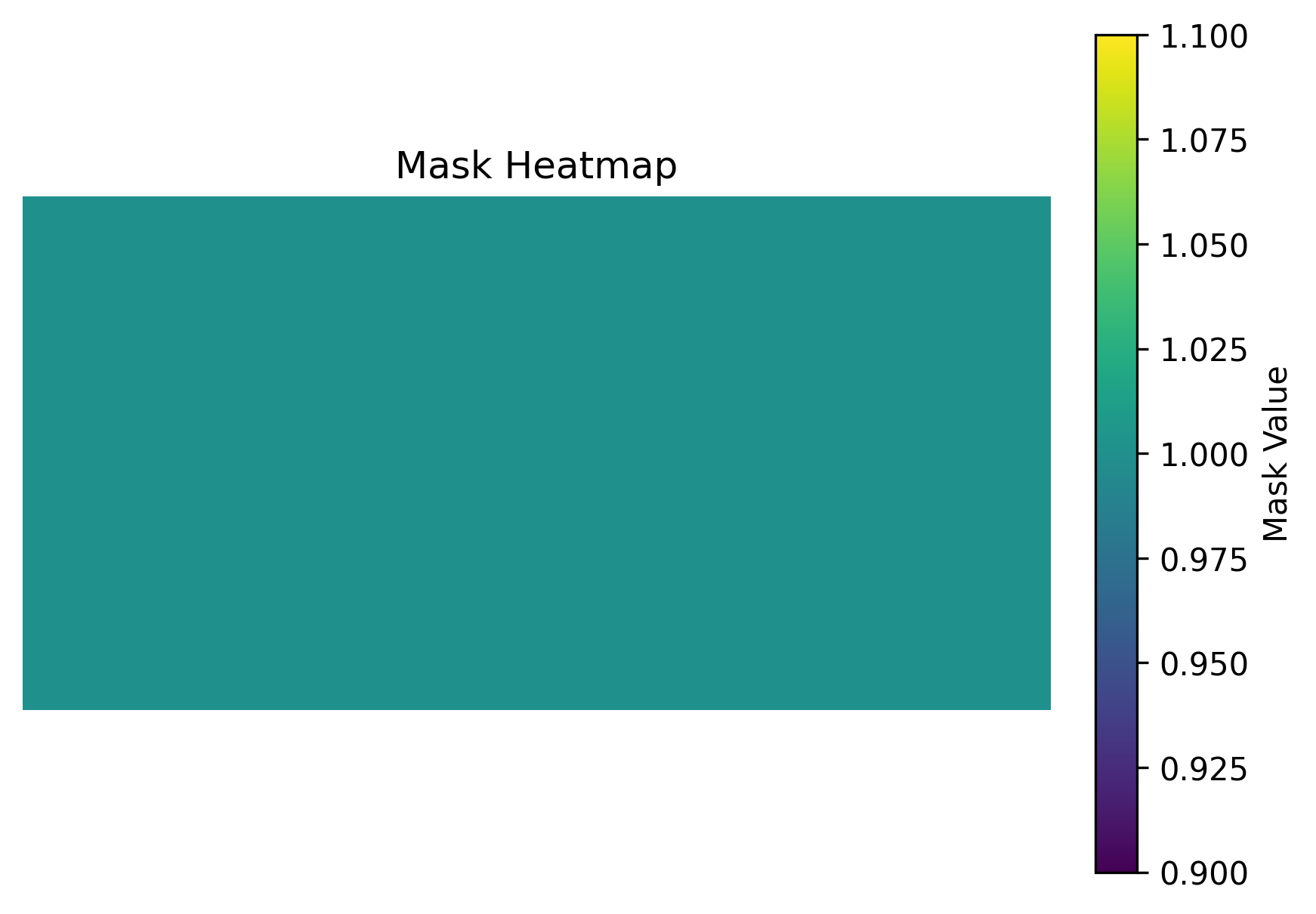}
    \caption{0.0031 bpp}
    \label{fig:noTS_low}
  \end{subfigure}
  \hfill
  \begin{subfigure}{0.48\linewidth}
    \centering
    \includegraphics[width=\linewidth]{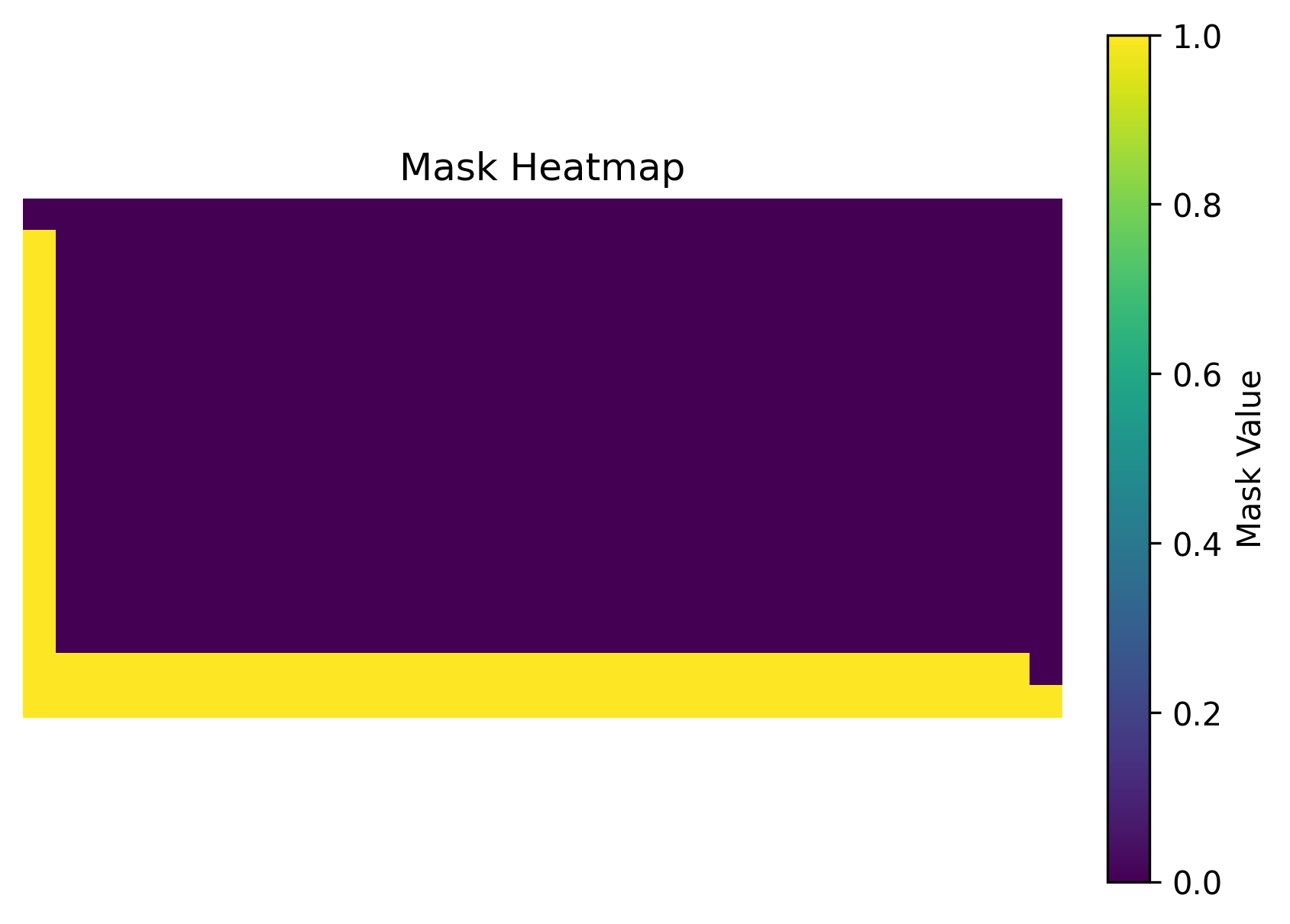}
    \caption{0.0554 bpp}
    \label{fig:noTS_high}
  \end{subfigure}

  \caption{Visualization of the visual feature mask under different bitrates without TS.}
  \label{fig:viswoTS}
\end{figure}
To visually illustrate the impact of textual side information on reconstruction quality, we further present local reconstruction comparisons, as illustrated in Fig.~\ref{ablationTSI}. In terms of fine-grained color reconstruction, the absence of textual side information tends to mix colors across adjacent regions, leading to inaccurate and less faithful color details. As shown in the Fig.~\ref{ablationTSI}, the boundary of the lane marking is contaminated by the shadow’s dark color, and the lack of certain color-transition cues results in local blurring around the window area. After incorporating textual side information, the diffusion model is guided to align its generative distribution with the specified global and local semantics, enabling it to better preserve structural details and produce more semantically accurate content.

Meanwhile, the textual side information also participates in the multimodal alignment auxiliary task, where it supervises the mask generator to focus on fine-grained details during optimization. As shown in Table~\ref{tsi}, removing the Textual Supervision (TS) severely degrades both perceptual and distortion metrics, with the perceptual metrics being particularly affected. Under the same LPIPS score, the absence of textual supervision increases the required bitrate by 121.49\%.

To provide a more intuitive comparison of the reconstruction quality, we further present corresponding visualizations, as illustrated in Fig.~\ref{ablationTS}. When the TS is removed, the reconstruction quality clearly deteriorates in fine-grained regions, exhibiting issues such as local blurring and boundary distortion. Moreover, as shown in Fig.~\ref{fig:viswoTS}, removing TS leads the generated masks to distribute their attention almost uniformly across the entire side-information map, rather than concentrating on specific regions. This indicates that a simple mask generator, without additional supervision, struggles to selectively filter local features, resulting in substantial redundant information and multi-view inconsistencies that ultimately 
degrade the reconstruction quality.

\section{Additional Results}

\subsection{Denoising Steps} 
\begin{figure}[htbp]
\centerline{\includegraphics[width=0.99\linewidth]{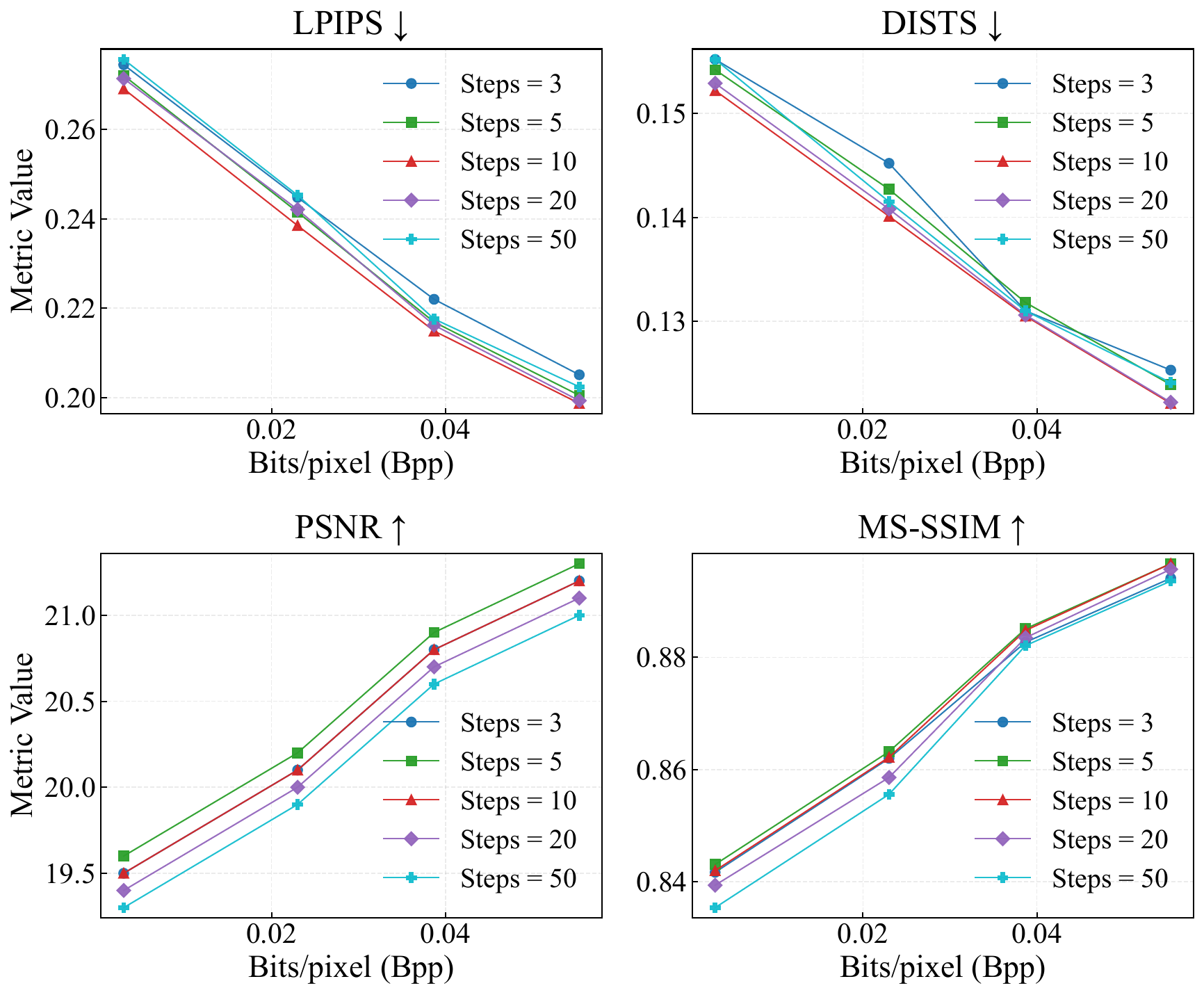}}
\caption{Evolution of LPIPS, DISTS, PSNR, and MS-SSIM across denoising steps as the number of different bitrates increases. Symbol conventions: $\uparrow$=higher better, $\downarrow$=lower better.}
\label{ds}
\end{figure}

As illustrated in Fig.~\ref{ds}, we evaluate the proposed method under different denoising steps using both perceptual and distortion metrics. The distortion metrics achieve their best performance when the denoising step is set to 5. As the number of steps increases to 10, the model attains the highest perceptual quality while maintaining nearly optimal MS-SSIM. However, when the denoising steps exceed 10, the overall reconstruction quality gradually declines. Therefore, all experiments in this paper adopt 10 denoising steps by default.

\subsection{Mask Visualization} 
\begin{figure}[htbp]
\centerline{\includegraphics[width=0.99\linewidth]{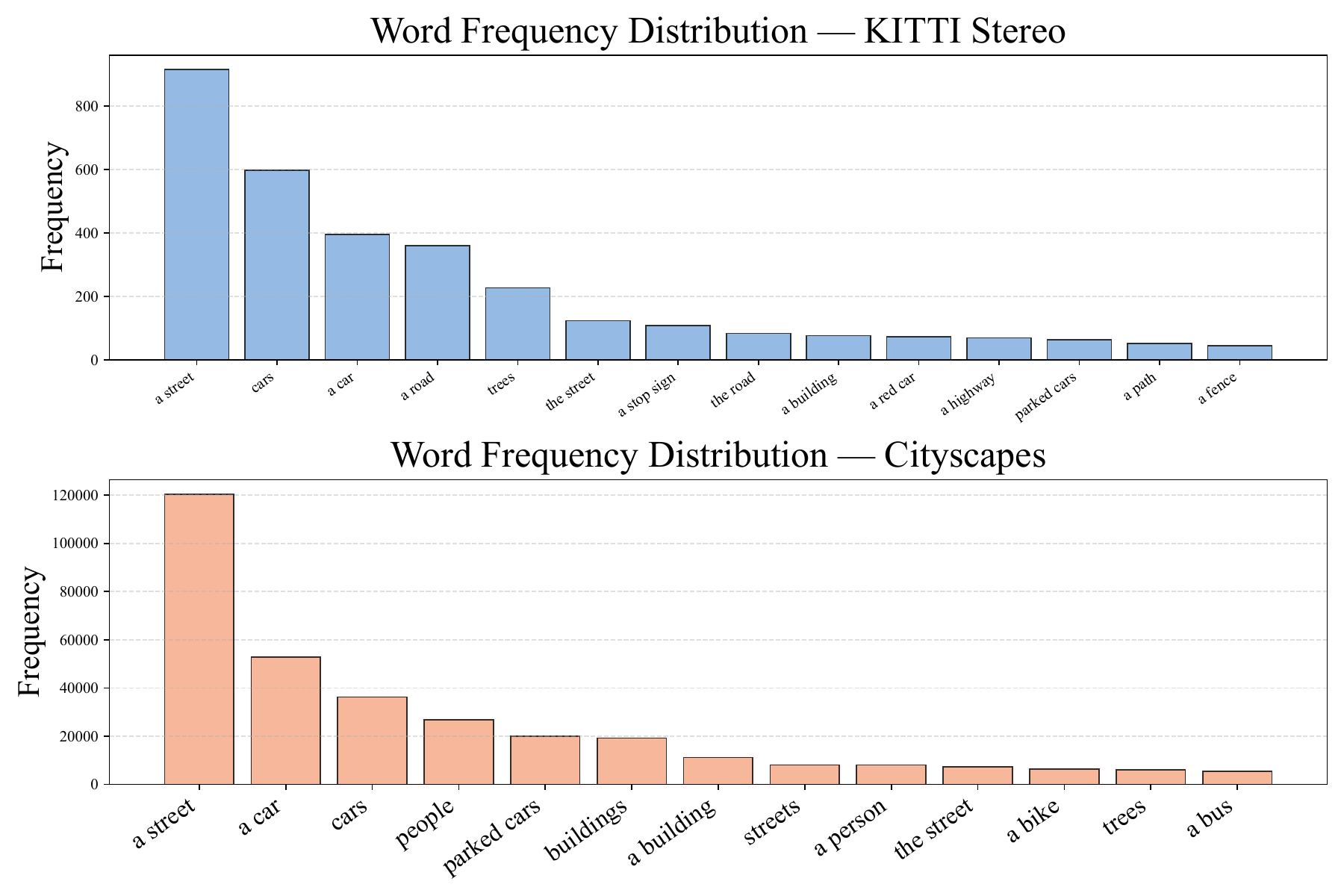}}
\caption{The high-frequency keywords on the KITTI Stereo and Cityscapes datasets. 
}
\label{word}
\end{figure}

We select the 14 most frequent object-level textual tokens from the KITTI Stereo and Cityscapes datasets to construct textual masks, and their frequency distributions are presented in Fig.~\ref{word}.

\begin{figure}[htbp]
\centerline{\includegraphics[width=0.99\linewidth]{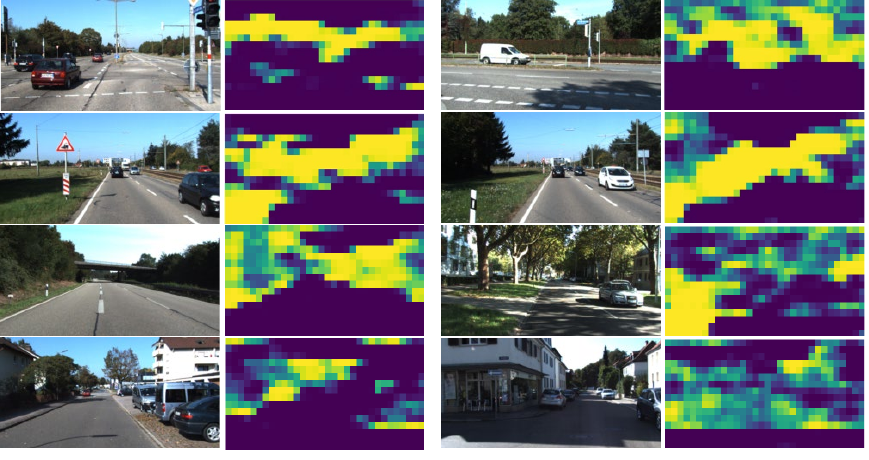}}
\caption{Visualization of the visual feature mask $m_v$. 
}
\label{mask}
\end{figure}

We visualize the corresponding visual masks in Fig.~\ref{mask}. With supervision from our object-level multimodal alignment task, the model learns to generate the visual mask conditioned on each input image, assigning different importance weights to different objects or regions. This helps address a key limitation in existing cross-attention–based methods, which struggle to effectively model global region-wise importance under extremely low bitrates.

\begin{table*}[htbp]
  \centering
    \caption{The comparison between our method and other diffusion-based approaches under different BD-Quality, as well as their encoding and decoding speeds. The BD-Quality measures the relative improvement or degradation of a given metric compared to a baseline at the same bitrate. DS represents the denoising steps.}
  \begin{tabular}{cccccccc}
    \hline
    \multirow{2}{*}{\textbf{Method}} & \multirow{2}{*}{\textbf{DS}} & \multicolumn{4}{c}{\textbf{BD-Quality $\uparrow$}} & \multicolumn{2}{c}{\textbf{Inference Time (s)}} \\
    \cline{3-8}
    & & \textbf{LPIPS} & \textbf{DISTS} & \textbf{PSNR} & \textbf{MS-SSIM} & \textbf{Encoding Time} & \textbf{Decoding Time} \\
    \hline
    \multirow{2}{*}{Perco~\cite{perco}} & 5 & -0.2501 & -0.1147 & -5.7031 & -5.8163 & $0.020\pm0.001$ & $0.320\pm0.002$ \\
    & 20 & -0.2364 & -0.1026 & -5.8774 & -5.8308 & $0.020\pm0.001$ & $0.843\pm0.002$ \\
    \hline
    \multirow{2}{*}{DiffEIC~\cite{diffeic}} & 20 & -0.2675 & -0.1000 & -5.7968 & -5.9150 & $0.114+0.005$ & $1.432\pm0.016$ \\
    & 50 & -0.2470 & -0.0931 & -5.9147 & -5.9622 & $0.114+0.005$ & $3.4371\pm0.025$ \\
    \hline
    \multirow{2}{*}{RDEIC~\cite{rdeic}} & 2 & -0.2295 & -0.1431 & -8.4241 & -4.4578 & $0.082\pm0.001$ & $0.121\pm0.001$ \\
    & 5 & -0.2300 & -0.1417 & -8.7294 & -4.5387 & $0.082\pm0.001$ & $0.295\pm0.002$ \\
    \hline
    \multirow{2}{*}{MDIC (Ours)} & 5 & -0.0040 & -0.0040 & 0.1144 & 0.0397 & $0.012\pm0.001$ & $0.461\pm0.002$ \\
    & 10 & 0 & 0 & 0 & 0 & $0.012\pm0.001$ & $0.638\pm0.006$ \\
    \hline
  \end{tabular}

  \label{complexity}
\end{table*}

As illustrated in Fig.~\ref{mask}, even among masked tokens, their importance is not identical. For example, the token "tree" receives consistently higher attention compared with other masked tokens such as "building" or "car". This is because trees typically contain richer structural and fine-grained texture information. Therefore, allocating more bit-budget to encode them is beneficial for reconstructing boundaries and color cues.

\begin{figure}[htbp]
\centerline{\includegraphics[width=0.99\linewidth]{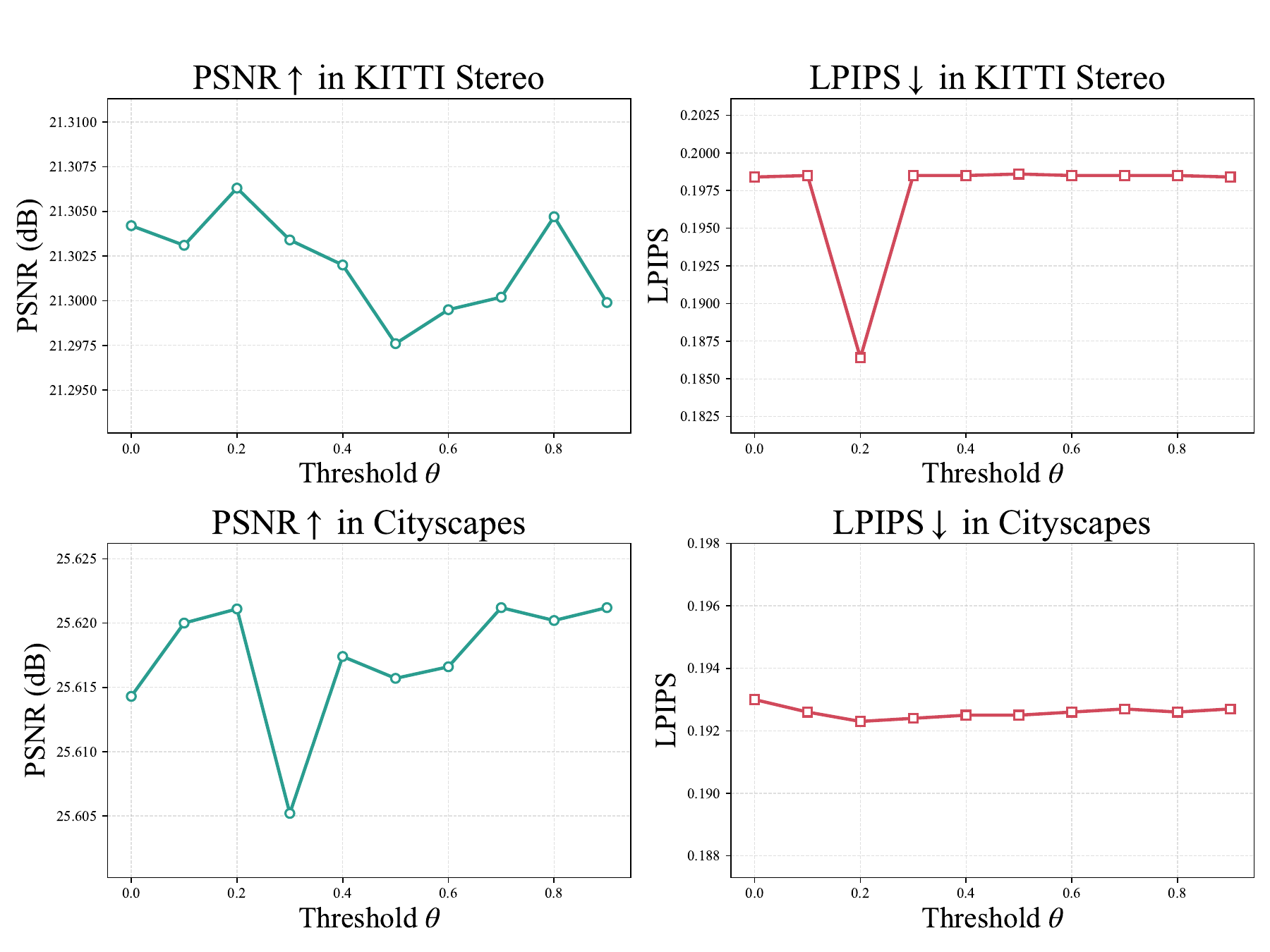}}
\caption{Variation of PSNR and LPIPS with threshold $\theta$ on the KITTI Stereo and Cityscapes datasets. 
}
\label{theta}
\end{figure}

To simultaneously select regions of varying importance and sparsify the continuous mask, we evaluate LPIPS and PSNR metrics under different thresholds $\theta$. For each threshold, metrics are averaged across all bitrates. As shown in Fig.~\ref{theta}, the best performance in both PSNR and LPIPS is achieved at a threshold of $\theta$=0.2.

\subsection{Generalizability and Robustness} 
\begin{figure}[htbp]
\centerline{\includegraphics[width=0.99\linewidth]{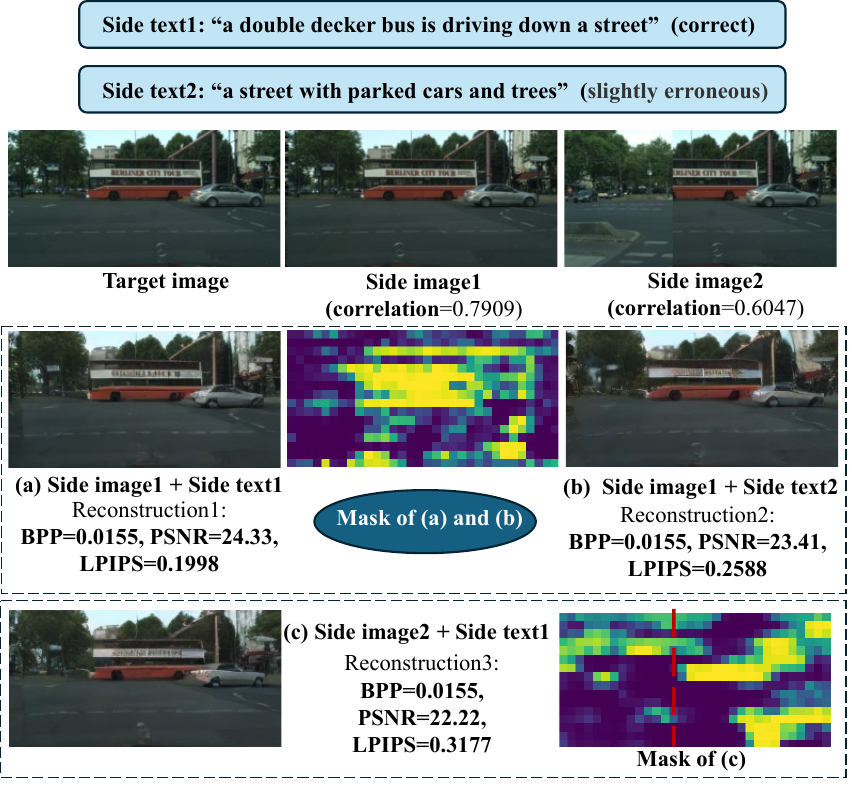}}
\caption{Applying the model trained on KITTI Stereo to the
Cityscapes dataset (unseen domain) without any fine-tuning. 
}
\label{re}
\end{figure}

As shown in Fig.~\ref{re}~(a), applying our KITTI-trained model directly to the unseen Cityscapes dataset yields high-quality reconstructions comparable to Fig.~\ref{qual} (main paper), verifying generalizability and robustness to domain shifts without fine-tuning. The reason is that the vocabulary acts strictly as training-time “Semantic Anchors” to supervise visual-semantic alignment, imposing no inference constraints. During inference, the mask generator relies solely on visual features to predict the mask without any text input. Therefore, it generalizes to unseen objects via learned structural features, akin to class-agnostic segmentation. As long as the vocabulary of size $N$ covers the test-time semantic space, the model generalizes without imposing any class-specific constraints during inference.

Furthermore, we use two side images with different degrees of correlation (measured by SSIM). In Fig.~\ref{re} (b) and (c), despite the expected quality degradation caused by incorrect captions or imperfect side information alignment, the model avoids collapse and continues to produce plausible images without severe artifacts. The mask successfully suppressed irrelevant regions (wrong road layout) and retained partial attention to the trees in the irrelevant region (sharing similar texture features) due to the latent similarity calculation and attention mechanism.

\subsection{Complexity Comparison} 
As shown in Table~\ref{complexity}, we compare our MDIC with representative diffusion-based image compression methods under different denoising steps in terms of Rate–Distortion-Perception (R-D-P) performance and inference speed.

For a fair comparison, we adopt the results of each method under its optimal denoising step as reported in the original papers. We then take our 10-step MDIC model as the baseline and compute the average metrics to measure relative performance variation at the same bitrate. Specifically, the BD-Quality is calculated by integrating the difference between the RD curves or RP curves of two methods over the overlapping bitrate range, where a positive value indicates improvement and a negative value denotes degradation compared with our baseline. The overall calculation process is as follows:
\begin{gather}
BD = \frac{1}{r_2 - r_1} \int_{r_1}^{r_2} [D_A(r) - D_B(r)] \, dr,
\end{gather}
where $D_A(r)$ and $D_B(r)$ denote the distortion or perception at bitrate $r$ for methods $A$ and $B$, respectively, and $[r_1, r_2]$ represents their common bitrate range.

In implementation, we fit the R-D or R-P points of each method using cubic interpolation and compute the definite integral of the curve differences over the overlapping bitrate range. This yields a unified and quantitative comparison across both perceptual and distortion metrics.

Diffusion-based LIC methods generally exhibit higher decoding complexity than VAE-based approaches due to their reliance on Stable Diffusion backbones. Although MDIC introduces an additional side-information interaction module compared to other diffusion-based LICs, it increases the inference time by only about 0.1s relative to Perco, while achieving notably better R-D-P performance at comparable bitrates.

\subsection{Qualitative Comparison} 
\begin{figure*}[!htb]
  \centering
  \setlength{\tabcolsep}{0.005\textwidth} 
  \begin{tabular}{*{6}{>{\centering\arraybackslash}p{0.159\textwidth}}} 
    \modeltag{figure/visual/k0_gt.png}{Ground Truth} & 
    \includegraphics[width=\linewidth]{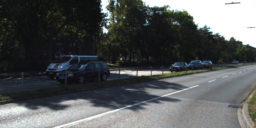} &
    \includegraphics[width=\linewidth]{figure/visual/k10_gt.png} &
    \includegraphics[width=\linewidth]{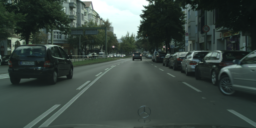} &
    \includegraphics[width=\linewidth]{figure/visual/c11_gt.png} &
    \includegraphics[width=\linewidth]{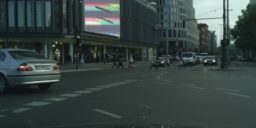} \\
    
    \multicolumn{6}{c}{}\\[-16pt] 
    
    \makebox[0.159\textwidth]{\scriptsize BPP, PSNR$\uparrow$, LPIPS$\downarrow$)} & 
    \makebox[0.159\textwidth]{\scriptsize (BPP, PSNR$\uparrow$, LPIPS$\downarrow$)} & 
    \makebox[0.159\textwidth]{\scriptsize (BPP, PSNR$\uparrow$, LPIPS$\downarrow$)} & 
    \makebox[0.159\textwidth]{\scriptsize (BPP, PSNR$\uparrow$, LPIPS$\downarrow$)} & 
    \makebox[0.159\textwidth]{\scriptsize (BPP, PSNR$\uparrow$, LPIPS$\downarrow$)} & 
    \makebox[0.159\textwidth]{\scriptsize (BPP, PSNR$\uparrow$, LPIPS$\downarrow$)} \\
    \multicolumn{6}{c}{}\\[-9pt] 


    \modeltag{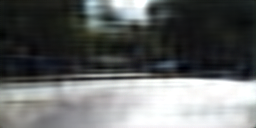}{NDIC} & 
    \includegraphics[width=\linewidth]{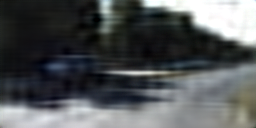} &
    \includegraphics[width=\linewidth]{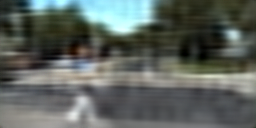} &
    \includegraphics[width=\linewidth]{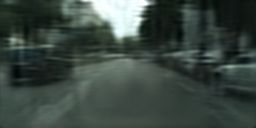} &
    \includegraphics[width=\linewidth]{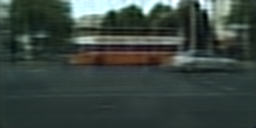} &
    \includegraphics[width=\linewidth]{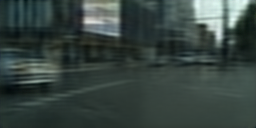} \\
    
    \multicolumn{6}{c}{}\\[-16pt]
    
    \makebox[0.159\textwidth]{\scriptsize (0.0265, 23.79, 0.4430)} & 
    \makebox[0.159\textwidth]{\scriptsize (0.0272, 25.04, 0.4171)} & 
    \makebox[0.159\textwidth]{\scriptsize (0.0280, 21.71, 0.4080)} & 
    \makebox[0.159\textwidth]{\scriptsize (0.0479, 26.03, 0.4207)} & 
    \makebox[0.159\textwidth]{\scriptsize (0.0478, 25.36, 0.3687)} & 
    \makebox[0.159\textwidth]{\scriptsize (0.0465, 26.21, 0.3561)} \\
    \multicolumn{6}{c}{}\\[-9pt] 

    \modeltag{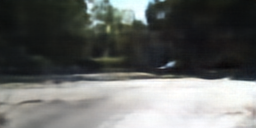}{LDMIC} & 
    \includegraphics[width=\linewidth]{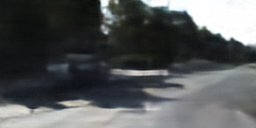} &
    \includegraphics[width=\linewidth]{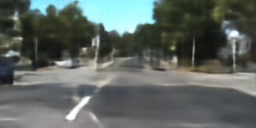} &
    \includegraphics[width=\linewidth]{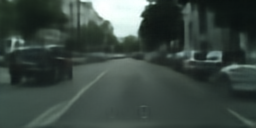} &
    \includegraphics[width=\linewidth]{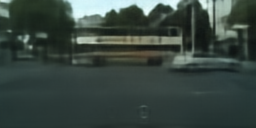} &
    \includegraphics[width=\linewidth]{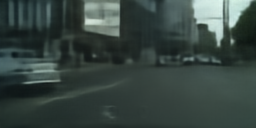} \\
    
    \multicolumn{6}{c}{}\\[-16pt]
    
    \makebox[0.159\textwidth]{\scriptsize (0.0143, 22.76, 0.4698)} & 
    \makebox[0.159\textwidth]{\scriptsize (0.0164, 24.00, 0.4497)} & 
    \makebox[0.159\textwidth]{\scriptsize (0.0288, 21.12, 0.4248)} & 
    \makebox[0.159\textwidth]{\scriptsize (0.0229, 27.43, 0.3101)} & 
    \makebox[0.159\textwidth]{\scriptsize (0.0281, 26.01, 0.3438)} & 
    \makebox[0.159\textwidth]{\scriptsize (0.0197, 26.96, 0.3449)} \\
    \multicolumn{6}{c}{}\\[-9pt] 

    \modeltag{figure/visual/k0_atn.png}{ATN} & 
    \includegraphics[width=\linewidth]{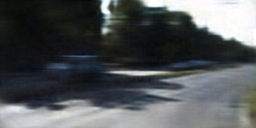} &
    \includegraphics[width=\linewidth]{figure/visual/k10_atn.png} &
    \includegraphics[width=\linewidth]{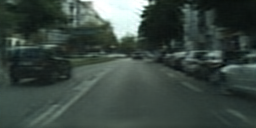} &
    \includegraphics[width=\linewidth]{figure/visual/c11_atn.png} &
    \includegraphics[width=\linewidth]{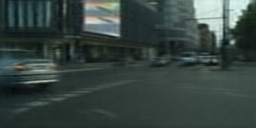} \\
    
    \multicolumn{6}{c}{}\\[-16pt]
    
    \makebox[0.159\textwidth]{\scriptsize (0.0316, 23.99, 0.4287)} & 
    \makebox[0.159\textwidth]{\scriptsize (0.0307, 22.51, 0.4150)} & 
    \makebox[0.159\textwidth]{\scriptsize (0.0401, 22.36, 0.3673)} & 
    \makebox[0.159\textwidth]{\scriptsize (0.0221, 28.96, 0.3000)} & 
    \makebox[0.159\textwidth]{\scriptsize (0.0248, 28.16, 0.2776)} & 
    \makebox[0.159\textwidth]{\scriptsize (0.0244, 28.52, 0.2540)} \\
    \multicolumn{6}{c}{}\\[-9pt] 

    \modeltag{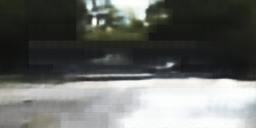}{ECSIC} & 
    \includegraphics[width=\linewidth]{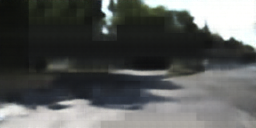} &
    \includegraphics[width=\linewidth]{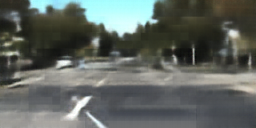} &
    \includegraphics[width=\linewidth]{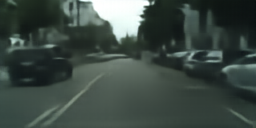} &
    \includegraphics[width=\linewidth]{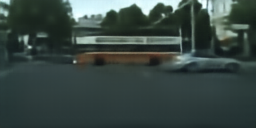} &
    \includegraphics[width=\linewidth]{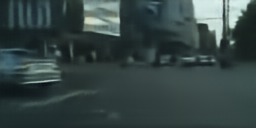} \\
    
    \multicolumn{6}{c}{}\\[-16pt]
    
    \makebox[0.159\textwidth]{\scriptsize (0.0342, 22.40, 0.4861)} & 
    \makebox[0.159\textwidth]{\scriptsize (0.0318, 23.64, 0.4572)} & 
    \makebox[0.159\textwidth]{\scriptsize (0.0397, 21.12, 0.4068)} & 
    \makebox[0.159\textwidth]{\scriptsize (0.0562, 26.95, 0.3642)} & 
    \makebox[0.159\textwidth]{\scriptsize (0.0490, 28.07, 0.3690)} & 
    \makebox[0.159\textwidth]{\scriptsize (0.0733, 25.98, 0.3243)} \\
    \multicolumn{6}{c}{}\\[-9pt] 

    \modeltag{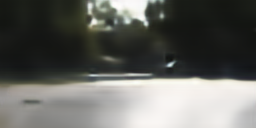}{SASIC} & 
    \includegraphics[width=\linewidth]{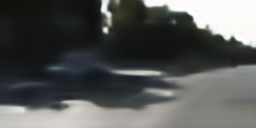} &
    \includegraphics[width=\linewidth]{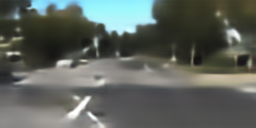} &
    \includegraphics[width=\linewidth]{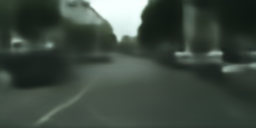} &
    \includegraphics[width=\linewidth]{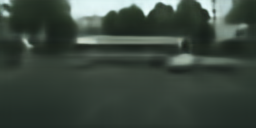} &
    \includegraphics[width=\linewidth]{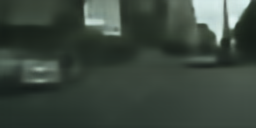} \\
    
    \multicolumn{6}{c}{}\\[-16pt]
    
    \makebox[0.159\textwidth]{\scriptsize (0.0268, 22.32, 0.5281)} & 
    \makebox[0.159\textwidth]{\scriptsize (0.0257, 23.52, 0.4783)} & 
    \makebox[0.159\textwidth]{\scriptsize (0.0397, 20.44, 0.4698)} & 
    \makebox[0.159\textwidth]{\scriptsize (0.0220, 23.40, 0.4858)} & 
    \makebox[0.159\textwidth]{\scriptsize (0.0202, 23.76, 0.4501)} & 
    \makebox[0.159\textwidth]{\scriptsize (0.0204, 24.47, 0.45311)} \\
    \multicolumn{6}{c}{}\\[-9pt] 

    \modeltag{figure/visual/k0_bisic.png}{BiSIC} & 
    \includegraphics[width=\linewidth]{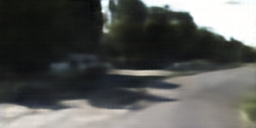} &
    \includegraphics[width=\linewidth]{figure/visual/k10_bisic.png} &
    \includegraphics[width=\linewidth]{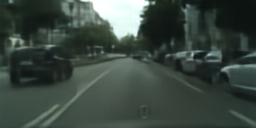} &
    \includegraphics[width=\linewidth]{figure/visual/c11_bisic.png} &
    \includegraphics[width=\linewidth]{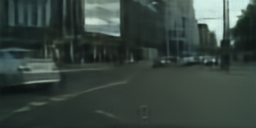} \\
    
    \multicolumn{6}{c}{}\\[-16pt]
    
    \makebox[0.159\textwidth]{\scriptsize (0.0288, 23.61, 0.4373)} & 
    \makebox[0.159\textwidth]{\scriptsize (0.0248, 24.83, 0.4253)} & 
    \makebox[0.159\textwidth]{\scriptsize (0.0408, 21.83, 0.4101)} & 
    \makebox[0.159\textwidth]{\scriptsize (0.0412, 28.78, 0.2492)} & 
    \makebox[0.159\textwidth]{\scriptsize (0.0460, 27.19, 0.3214)} & 
    \makebox[0.159\textwidth]{\scriptsize (0.0404, 28.17, 0.2699)} \\
    \multicolumn{6}{c}{}\\[-9pt] 

    \modeltag{figure/visual/k0_perco.png}{Perco} & 
    \includegraphics[width=\linewidth]{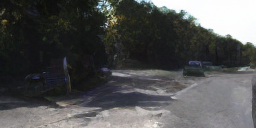} &
    \includegraphics[width=\linewidth]{figure/visual/k10_perco.png} &
    \includegraphics[width=\linewidth]{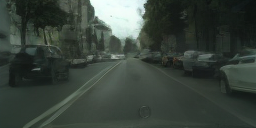} &
    \includegraphics[width=\linewidth]{figure/visual/c11_perco.png} &
    \includegraphics[width=\linewidth]{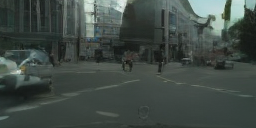} \\
    
    \multicolumn{6}{c}{}\\[-16pt]
    
    \makebox[0.159\textwidth]{\scriptsize (0.0313, 17.85, 0.4385)} & 
    \makebox[0.159\textwidth]{\scriptsize (0.0313, 17.90, 0.4212)} & 
    \makebox[0.159\textwidth]{\scriptsize (0.0313, 15.16, 0.4171)} & 
    \makebox[0.159\textwidth]{\scriptsize (0.0313, 20.57, 0.3257)} & 
    \makebox[0.159\textwidth]{\scriptsize (0.0313, 18.17, 0.3777)} & 
    \makebox[0.159\textwidth]{\scriptsize (0.0313, 19.01, 0.3557)} \\
    \multicolumn{6}{c}{}\\[-9pt] 

    \modeltag{figure/visual/k0_rdeic.png}{RDEIC} & 
    \includegraphics[width=\linewidth]{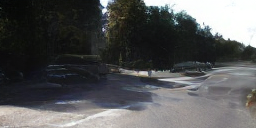} &
    \includegraphics[width=\linewidth]{figure/visual/k10_rdeic.png} &
    \includegraphics[width=\linewidth]{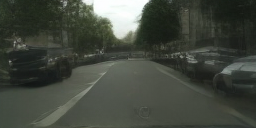} &
    \includegraphics[width=\linewidth]{figure/visual/c11_rdeic.png} &
    \includegraphics[width=\linewidth]{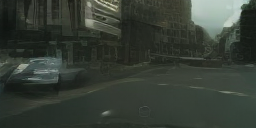} \\
    
    \multicolumn{6}{c}{}\\[-16pt]
    
    \makebox[0.159\textwidth]{\scriptsize (0.0188, 12.71, 0.5445)} & 
    \makebox[0.159\textwidth]{\scriptsize (0.0207, 12.06, 0.5530)} & 
    \makebox[0.159\textwidth]{\scriptsize (0.0212, 12.61, 0.5660)} & 
    \makebox[0.159\textwidth]{\scriptsize (0.0373, 20.96, 0.3668)} & 
    \makebox[0.159\textwidth]{\scriptsize (0.0425, 20.21, 0.3888)} & 
    \makebox[0.159\textwidth]{\scriptsize (0.0405, 20.37, 0.4401)} \\
    \multicolumn{6}{c}{}\\[-9pt] 

    \modeltag{figure/visual/k0_fgpdic.png}{MDIC(Ours)} & 
    \includegraphics[width=\linewidth]{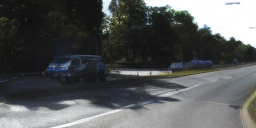} &
    \includegraphics[width=\linewidth]{figure/visual/k10_fgpdic.png} &
    \includegraphics[width=\linewidth]{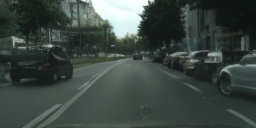} &
    \includegraphics[width=\linewidth]{figure/visual/c11_fgpdic.png} &
    \includegraphics[width=\linewidth]{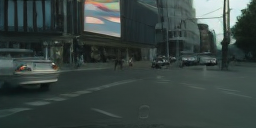} \\
    
    \multicolumn{6}{c}{}\\[-16pt]
    
    \makebox[0.159\textwidth]{\scriptsize (0.0107, 22.56, 0.2524)} & 
    \makebox[0.159\textwidth]{\scriptsize (0.0134, 24.20, 0.2196)} & 
    \makebox[0.159\textwidth]{\scriptsize (0.0096, 20.61, 0.2391)} & 
    \makebox[0.159\textwidth]{\scriptsize (0.0186, 26.13, 0.1551)} & 
    \makebox[0.159\textwidth]{\scriptsize (0.0207, 24.61, 0.1663)} & 
    \makebox[0.159\textwidth]{\scriptsize (0.0201, 25.71, 0.1663)} \\
    \multicolumn{6}{c}{}\\[-9pt]

    \makebox[0.159\textwidth]{\scriptsize (a)} & 
    \makebox[0.159\textwidth]{\scriptsize (b)} & 
    \makebox[0.159\textwidth]{\scriptsize (c)} & 
    \makebox[0.159\textwidth]{\scriptsize (d)} & 
    \makebox[0.159\textwidth]{\scriptsize (e)} & 
    \makebox[0.159\textwidth]{\scriptsize (f)}
  \end{tabular}
  
  \caption{
    Visualization results on KITTI Stereo and Cityscapes datasets. Each row corresponds to a different method. 
    Numbers below the images indicate (BPP, PSNR, LPIPS). Symbol conventions: $\uparrow$=higher better, $\downarrow$=lower better. 
}
  
  \label{qual1}
\end{figure*}

As shown in Fig.~\ref{qual1}, we visualize the reconstruction results of all compared methods under approximately the same compression rate.
Existing distortion-oriented DIC methods (NDIC~\cite{ndic}, ATN~\cite{atn}, LDMIC~\cite{ldmic}) achieve high pixel-level fidelity and maintain good semantic consistency with the original images. However, they still suffer from noticeable blurring and artifacts in both local and global regions.
Joint encoding methods (SASIC~\cite{sasic}, ECSIC~\cite{ecsic}, BiSIC~\cite{bisic}) enable additional interaction among multi-view images during the encoding stage, thereby improving the overall perceptual quality compared to conventional DIC approaches. Nonetheless, due to the extremely low bitrate and the limited amount of transmitted information, these methods lose significant fine-grained details, leading to local blurriness and texture degradation.
Perception-oriented LIC methods (Perco~\cite{perco}, DiffEIC~\cite{diffeic}, RDEIC~\cite{rdeic}) demonstrate better perceptual results in both global and local aspects. However, the absence of semantic consistency constraints and the lack of effective side-information utilization often result in semantically inconsistent details and distorted edge structures that deviate from the original image content.
In contrast, the proposed MDIC achieves the best global and local perceptual quality. Even under extremely low bitrates, it maintains high perceptual fidelity and fine-grained semantic consistency with original image, clearly demonstrating the superiority of MDIC in effectively leveraging side information for reconstruction.

\end{document}